\documentclass[10pt,twocolumn,letterpaper]{article}
\pdfoutput=1

\usepackage{cvpr}
\usepackage{times}
\usepackage{epsfig}
\usepackage{graphicx}
\usepackage{amsmath}
\usepackage{amssymb}

\usepackage{titling}

\usepackage[pagebackref=true,breaklinks=true,letterpaper=true,colorlinks,bookmarks=false]{hyperref}

\usepackage{xcolor}
\usepackage{animate}
\usepackage{multirow}
\usepackage{arydshln} 
\usepackage{enumitem} 
\usepackage{cite} 

\cvprfinalcopy 

\def\cvprPaperID{-} 

\newcommand{\animbox}[1]{\textcolor{red}{\fboxrule=0.1pt,\fboxsep=0.1pt\fbox{#1}}}

\newcommand{\Ishort}{\ensuremath{Y_{0^-}}}
\newcommand{\Ishortt}{\ensuremath{Y_{1^+}}}
\newcommand{\Iblur}{\ensuremath{Y_{0 \rightarrow 1}}}

\newcommand{\Ihalfgt}{\ensuremath{{I}_{0 \rightarrow 0.5{^-}}}}
\newcommand{\Ihalffgt}{\ensuremath{{I}_{0.5{^+} \rightarrow 1}}}
\newcommand{\Isharpgt}{\ensuremath{{I}_{0.5}}}

\newcommand{\Ihalf}{\ensuremath{\widehat{I}_{0 \rightarrow 0.5{^-}}}}
\newcommand{\Ihalff}{\ensuremath{\widehat{I}_{0.5{^+} \rightarrow 1}}}
\newcommand{\Isharp}{\ensuremath{\widehat{I}_{0.5}}}

\newcommand{\Isharpx}{\ensuremath{\widehat{I}_{0.25}}}
\newcommand{\Isharpy}{\ensuremath{\widehat{I}_{0.75}}}

\newcommand{\Iquarter}{\ensuremath{\widehat{I}_{0 \rightarrow 0.25{^-}}}}
\newcommand{\Iquarterr}{\ensuremath{\widehat{I}_{0.25{^+} \rightarrow 0.5{^-}}}}
\newcommand{\Iquarterrr}{\ensuremath{\widehat{I}_{0.5{^+} \rightarrow 0.75{^-}}}}
\newcommand{\Iquarterrrr}{\ensuremath{\widehat{I}_{0.75{^+} \rightarrow 1}}}

\ifcvprfinal\fi
\begin{document}

\title{\vspace*{-2cm}Photosequencing of Motion Blur using Short and Long Exposures}

\author{Vijay Rengarajan\\
Carnegie Mellon University\\
{\tt\small vangarai@andrew.cmu.edu}
\and
Shuo Zhao\\
Carnegie Mellon University\\
{\tt\small shuozhao1991@gmail.com}
\and
Ruiwen Zhen\\
Samsung Research America\\
{\tt\small r.zhen@samsung.com}
\and
John Glotzbach\\
Samsung Research America\\
{\tt\small j.glotzbach@samsung.com}
\and
Hamid Sheikh\\
Samsung Research America\\
{\tt\small hr.sheikh@samsung.com}
\and
Aswin C.~Sankaranarayanan\\
Carnegie Mellon University\\
{\tt\small saswin@andrew.cmu.edu}
}

\date{}
\maketitle
\ \\[-1.25cm]
\begin{abstract}
Photosequencing aims to transform a motion blurred image to a sequence of sharp images.
This problem is challenging due to the inherent ambiguities in temporal ordering as well as the recovery of lost spatial textures due to blur. 
Adopting a computational photography approach, we propose to capture two short exposure images, along with the original blurred long exposure image to aid in the aforementioned challenges.
Post-capture, we recover the sharp photosequence using a novel blur decomposition strategy that recursively splits the long exposure image into smaller exposure intervals. 
We validate the approach by capturing a variety of scenes with interesting motions using machine vision cameras programmed to capture short and long exposure sequences.
Our experimental results show that the proposed method resolves both fast and fine motions better than prior works.
\end{abstract}

\section{Introduction}
Capturing photosequences of fast events is a challenge in the photography milieu. High speed capture is costly to implement since it requires specialized electronics capable of handling the high bandwidth of data as well as highly sensitive image sensors that can make low noise measurements at short exposures. As a consequence, it is still a major obstacle in commodity cameras, especially without sacrificing the full spatial  resolution of the sensor. This paper provides a pathway for the implementation of such a capability with no additional hardware. 

\begin{figure}[t]
    \begin{center}
    \renewcommand{\tabcolsep}{0.5pt}
    \small
    \begin{tabular}{cccc}
    \multicolumn{4}{c}{\!\!\includegraphics[width=\linewidth]{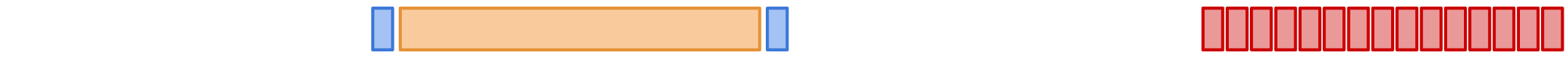}}\\
    \includegraphics[width=0.24\linewidth]{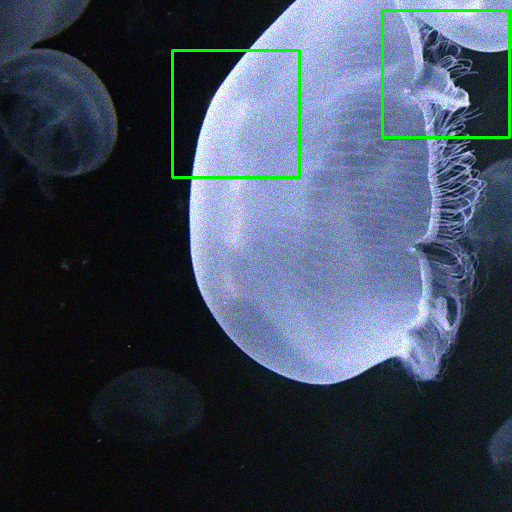}&
    \includegraphics[width=0.24\linewidth]{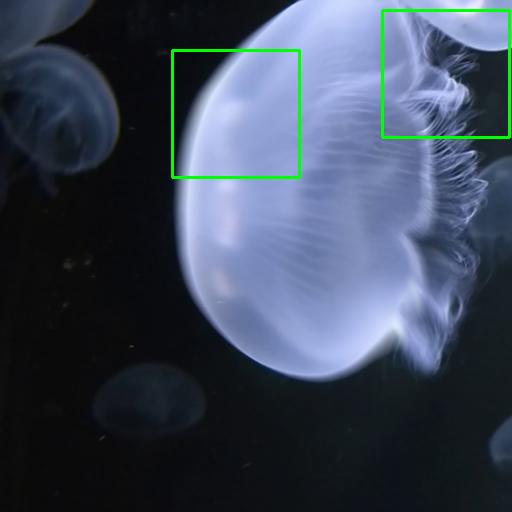}&
    \includegraphics[width=0.24\linewidth]{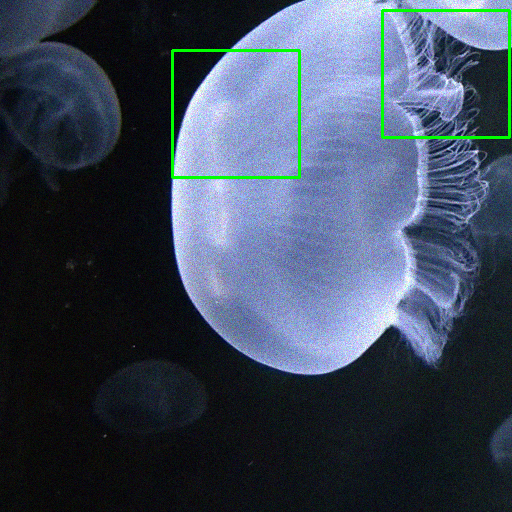}&
    \animbox{\animategraphics[width=0.24\linewidth,loop]{5}{img/intro/jellyfish3/c}{01}{15}}\\
    \includegraphics[width=0.24\linewidth]{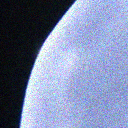}&
    \includegraphics[width=0.24\linewidth]{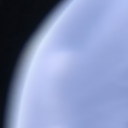}&
    \includegraphics[width=0.24\linewidth]{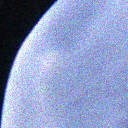}&
    \animbox{\animategraphics[width=0.24\linewidth,loop]{5}{img/intro/jellyfish3/crop5/c}{01}{15}}\\
    \\
    \includegraphics[width=0.24\linewidth]{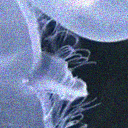}&
    \includegraphics[width=0.24\linewidth]{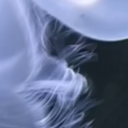}&
    \includegraphics[width=0.24\linewidth]{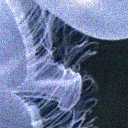}&
    \animbox{\animategraphics[width=0.24\linewidth,loop]{5}{img/intro/jellyfish3/crop6/c}{01}{15}}\\
    (a) Short exp.& (b) Long  exp.& (c) Short exp.  & (d) Photoseq.\\
    \multicolumn{4}{c}{Click the images in (d) to play animation in Adobe Reader\textsuperscript{\textcopyright}.}
    \end{tabular}
    \end{center}
    \caption{We capture three photographs in quick succession, a short-long-short 1ms-15ms-1ms exposure triplet (a-c), to capture both  textures and fast motions. The short exposures (a,c) capture sharpness while being noisy. The long exposure (b) captures motion in the form of blur. (d) Our photosequencing method leverages the complementary information from these exposures to recover the s scene with smooth motion and sharp textures.}
    \label{fig:intro}
 \end{figure}

One approach is to recover photosequences from motion blur. A single motion-blurred photograph embeds motion events inherently, but recovering them can be hard due to ill-posed temporal ordering, as well as the erasure of spatial textures. While prior single-blur sequencing approaches~\cite{jin:2018:learning, purohit:2018:bringing} handle both these issues using data-driven priors encoded using deep neural networks, they are incapable of correctly sequencing multiple local motion events happening across the sensor space. A recent multi-blur sequencing technique~\cite{jin:2019:learning} handles the overall sequencing problem by processing multiple blurred photographs captured in succession. As much as texture recovery from blur is conditioned using deep neural networks, fine textures can still be lost due to fast motion since all of the captured photographs are blurred. An orthogonal approach to motion blur is to first capture a short-exposure photosequence at higher frame-rates followed by frame-wise denoising~\cite{suo:2014:joint}. Even though this looks promising, texture loss could still be an issue due to denoising. Further, static  regions, which would have benefited from long exposures,  suffer  due to postprocessing which would otherwise be clean.

This paper proposes to capture a short-long-short exposure triplet to handle both motion ordering and texture recovery in photosequencing. The two additional short-exposure photographs are designed to have an exposure that is much smaller than the original long exposure photograph. Hence, while these additional images might be noisy, they are practically free of motion-blur. Further, the short exposure photographs resolve temporal ordering by acting as peripheral keypoints of the long exposure. 
Such a capture mechanism is easily implemented in modern cameras since they are endowed with a burst mode.
Once we capture the short-long-short exposure images, we computationally recover the photosequence equivalent to a high-speed capture of the underlying scene. The blurred image is first decomposed into two \textit{half-blurred} images with reduced blur corresponding to the two halves of the exposure period and a single sharp image corresponding to the mid-point. We learn this decomposition by training a deep neural network using images synthesized from high frame-rate video datasets. A recursive decomposition leads to the sharp photosequence without any enforced temporal order. The recovered images gain from the complementary goodness of the short and long exposures for static and moving scene regions.

Figs.~\ref{fig:intro}(a,b,c) show the short-long-short exposure images captured in succession of a scene with a moving jellyfish. The short exposures are noisy while the long exposure embeds motion blur. The intricate tentacles are captured in the noisy image but are blurred in the long exposure. Our recovered 15x photosequence corresponding to the long exposure duration is shown in (d). Both texture and motion are recovered correctly in our reconstruction.

The main contributions of this work are as follows:
\begin{enumerate}[leftmargin=*]
\itemsep0em 
    \item We propose the novel idea of capturing short-long-short exposure images for the problem of photosequencing of motion blur. 
    \item We propose a recursive blur decomposition strategy in which increase in temporal super-resolution can be achieved through multiple levels of decomposition till the removal of blur. 
    \item We present a new short-long-short exposure dataset for a variety of dynamic events captured using a machine vision camera. We provide qualitative evaluation and comparisons to prior art on this data.
\end{enumerate}

\section{Related Works}
Traditional deblurring approaches \cite{chan:1998:total,krishnan:2011:blind,pan:2016:blind,zhang:2018:dynamic,tao:2018:scale, chakrabarti:2016:neural,sun:2015:learning} aim to get a single sharp image along with local motion kernels and camera trajectories. 
Reconstructing the whole photosequence for combined object and camera motions has been dealt only recently~\cite{jin:2018:learning,purohit:2018:bringing}. Single-blur techniques produce photosequence from a single image by imposing ordering constraints through explicit temporal image costs as in \cite{jin:2018:learning} or through an implicit cost using a video encoder-decoder framework as in \cite{purohit:2018:bringing}. These methods suffer from ambiguities in temporal ordering of multiple objects. The recent multi-blur approach \cite{jin:2019:learning} does photosequencing using multiple motion-blurred images to impart more information for reconstruction and ordering. All these photosequencing approaches use neural network priors to reconstruct sharp textures from blur. 

One can also increase framerate through temporal interpolation of video frames. An interpolation approach based on gradients and phase is employed in \cite{mahajan:2009:moving} and \cite{meyer:2015:phase}, respectively, while \cite{shahar:2011:space} fuses recurring video patches to perform spatio-temporal super-resolution. Recently, \cite{jiang:2018:super} employed neural networks to linearly interpolate optical flows. However, frame interpolation methods are inherently limited from the assumption of video frames being non-blurred. 

Pre-regularizing the ill-posedness of image recovery during capture time is a central tenet of computational photography. 
A hybrid high resolution, lower frame rate and a low resolution, higher frame rate camera setup is used in \cite{ben:2003:motion,tai:2008:image} to remove motion blur from the high resolution capture, but a further photosequencing step is not performed. The coded exposure work~\cite{raskar:2006:coded} suggests changing exposure design to improve the conditioning of the deblurring problem. The idea of capturing a noisy-blurry pair in \cite{yuan:2007:image} leads to the extraction of complementary information that regularizes the deblurring problem better. While this work employs an additional short exposure image to produce a single sharp image, our goal is to recover the whole photosequence.

\begin{figure*}
    \begin{center}
    \includegraphics[width=0.99\linewidth]{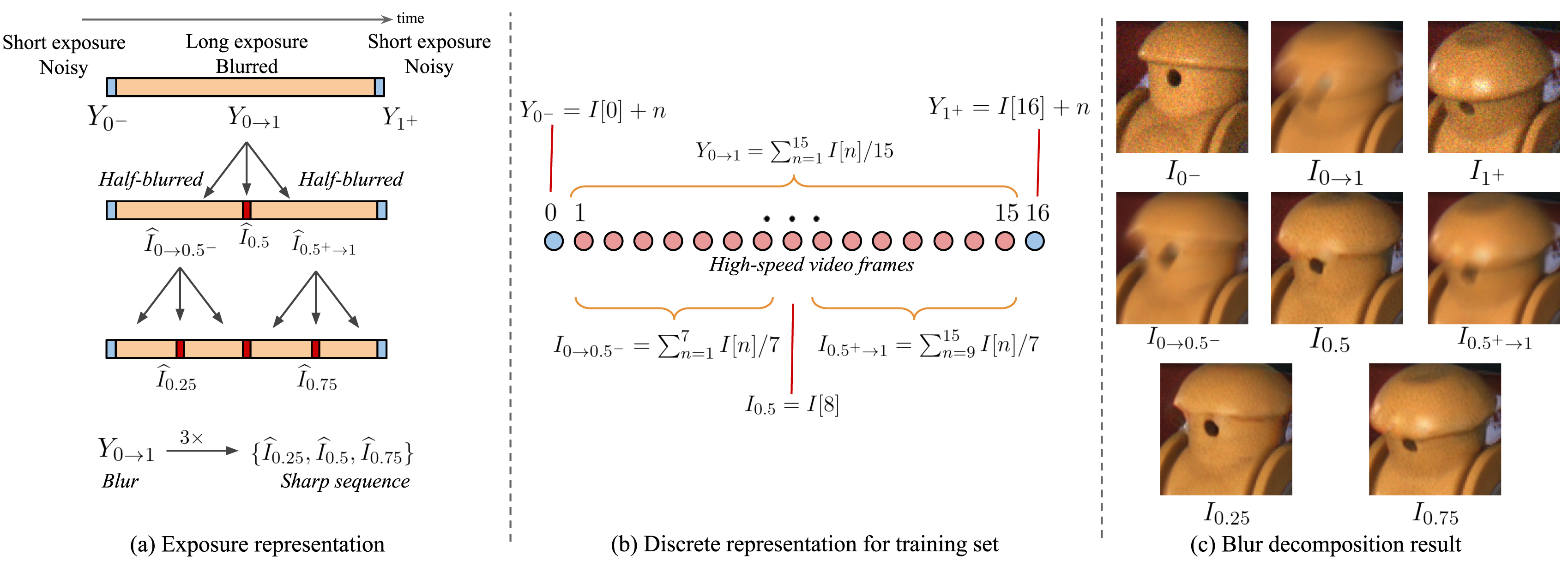}
    \end{center}
    \caption{\textit{Recursive blur decomposition till 3$\times$ recovery.} Our method takes short-long-short exposure triplet as inputs to produce a sharp sequence of images. Our core decomposition step is to split the blurred image into two half-blurred images and a midpoint sharp image. On recursive decomposition, we arrive at the desired sequence of sharp images. Two levels of decomposition, i.e. 3x recovery, is illustrated in (a). The discrete representation for preparing training set for an example of blurring 15 images is illustrated in (b). In (c), we show an example of our blur decomposition.} 
    \label{fig:blur_decomp}
\end{figure*}

\section{Problem}
Let $I_t$ be the clean ground-truth image of a scene at $t \in [0,1]$. Let $Y(a,b)$ be an observed image for the time interval $[a,b]$ defined as $Y(a,b) = 1/(b-a)~.~\int_{a}^{b} I_t dt +n$ where $n$ represents noise. The goal of motion-blur photosequencing is capture a blurred image $\Iblur = Y(0,1)$ and to estimate the images $\{\widehat{I}_t{_j}\}_{j=1}^N$ such that $\Iblur \approx \sum_{j=1}^N \widehat{I}_t{_j}/N$ where $t_j$s are equi-spaced timepoints and $N$ represents the \textit{sequence-rate}. In essence, the recovered  sequence represents the series of images of the scene if it had been captured with a hypothetical higher frame-rate camera operating at the same spatial resolution and without the noise level associated with that frame rate.

\subsection{Approach}
We propose to capture two short exposure images $\Ishort = Y(-\Delta t, 0)$ and $\Ishortt = Y(1,1+\Delta t)$, one before and one after the long exposure image $\Iblur$. Though the short exposure images will be noisy owing to the very short exposure interval $\Delta t$, they provide the much-needed information of scene texture. They also act as temporal endpoints for the blurred image to disambiguate motion directions. 
Our problem statement is to estimate the image sequence $\{\widehat{I}_{t{_j}}\}$ happened during long exposure, given the three input images, $\{\Ishort, \Iblur, \Ishortt\}$ as inputs. 

\subsection{Recursive Blur Decomposition}
\label{sec:blur_decomp}
We adopt a multi-step sequencing strategy wherein we progressively increase the number of reconstructed sharp images. We first decompose the long exposure into two half-blurred images $\Ihalf$ and $\Ihalff$ and the sharp image $\Isharp$. 
Estimating just $\Isharp$ would correspond to deblurring. In addition, we also estimate $\Ihalf$ and $\Ihalff$ that contain lesser motion blur corresponding to each half of the original exposure interval. Our core blur decomposition step, hence, is the following:
\begin{align} 
    \{\Ishort, \Iblur, \Ishortt\} \rightarrow  \{\Ihalf, \Isharp, \Ihalff\}.
    \label{eq:blur_decomp}
 \end{align}

Our idea then is to perform blur decomposition on both the half-blurred images to further split the blur interval and get a sharp image corresponding to their respective middle timepoints as shown in Fig.~\ref{fig:blur_decomp}(a). 
The second-level will result in the sharp sequence $\Isharpx, \Isharp, \Isharpy$ corresponding to 3x frame-rate. 
We could perform blur decomposition recursively to achieve a desired sequence-rate. For instance, a third level of blur decomposition would provide us with 7x sequence-rate, a fourth level 15x, and so on. In practice, one could stop the decomposition at a desired level when the blur present in half-blurred images is negligible. 

\section{Implementation}
We learn the blur decomposition mapping in \eqref{eq:blur_decomp} by training a neural network.

\subsection{Network Architecture}
\begin{figure}
\begin{center}
    \includegraphics[width=0.99\linewidth]{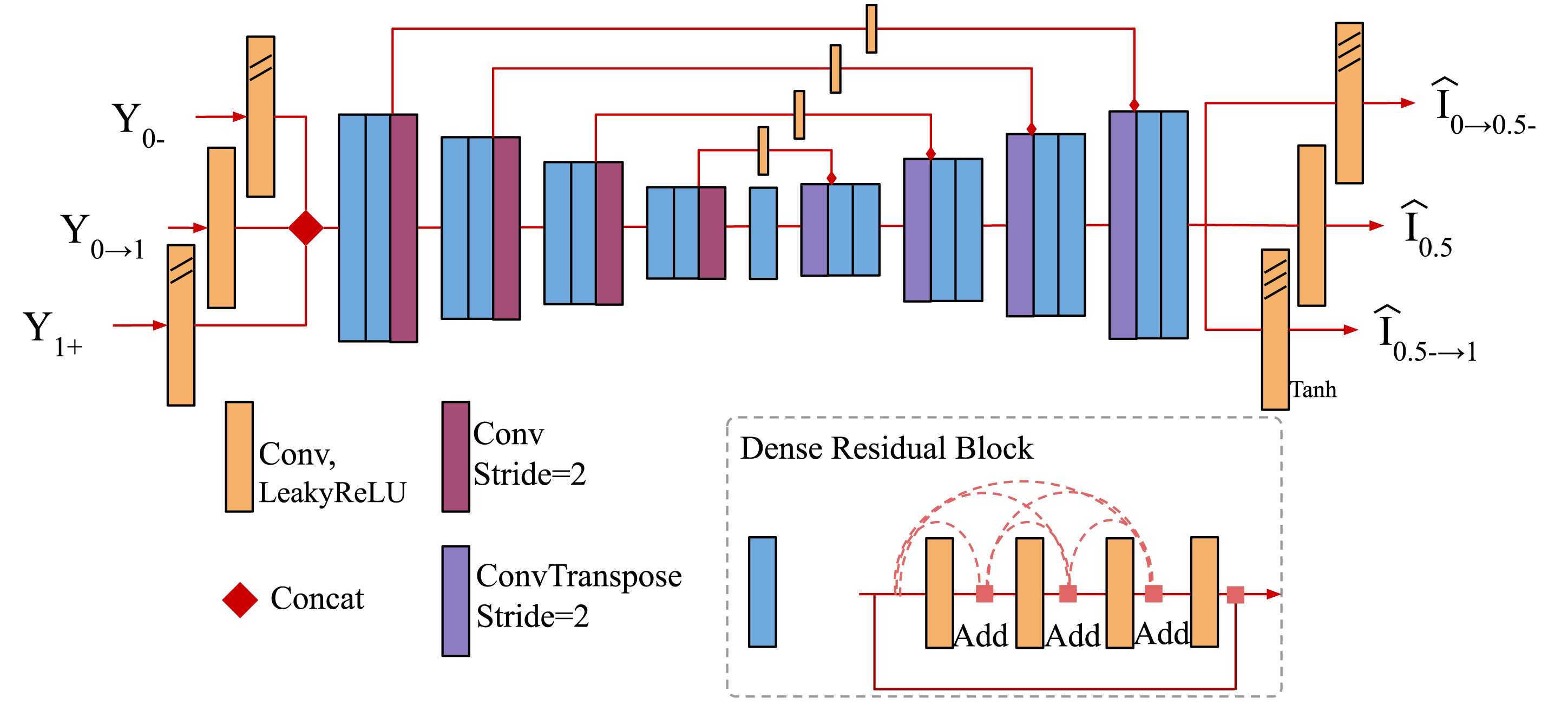}
\end{center}
\caption{\textit{Blur decomposition network.} We use a Unet-like architecture with dense residual blocks to provide rich local connections for encoder feature extraction and carry-on convolutions for encoder-to-decoder feature transfers.}
\label{fig:arch}
\end{figure}

We adopt an encoder-decoder architecture similar to U-net \cite{ronneberger:2015:unet, isola:2017:image} as shown in Fig.~\ref{fig:arch}. We use dense residual blocks~\cite{zhang:2018:residual} which have rich local connections, instead of serialized convolutional layers. We use carry-on convolutional layers through the skip connections from encoder to decoder. For our network, the inputs are the short-long-short exposure observations $\{\Ishort, \Iblur, \Ishortt\}$ 
and the outputs are the estimates of half-blurred images and the deblurred image, $\{\Ihalf, \Isharp, \Ihalff\}$. The input images pass through different initial convolutional layers; similarly the output layers are different for half-blurred and the deblurred images. The input layer of the short exposures share the same weights; the output layer of the half-blurred images share the same weights.
The image intensities are in the range [0,1]. All the convolutional layers are followed by Leaky ReLU nonlinearity except for the last layer which is followed by rescaled-tanh to enforce the output range to [0,1]. All convolutional layers use 3x3 filters except for the first layers of both noisy and blurred images which use a filter size of 7x7. 
More details of the architecture are provided in the supplementary material.

\subsection{Training Data}
\label{sec:train_data}
Our goal is to have the neural network learn to decompose different amounts of blur in the long exposure image with the help of sharp and noisy short exposure images. 
Training a neural network requires a large number of blurred, half-blurred, and sharp images according to \eqref{eq:blur_decomp}. Hence, we use existing high-speed 240fps video datasets to create our training set. We employ multiple video datasets, Adobe240fps \cite{su:2017:deep}, GoProTrain240fps \cite{nah:2017:deep}, and Sony240fps \cite{jin:2019:learning}, to avoid camera bias. 
Our single training sample is defined by random 128x128 crops created from full-sized video frames.
We follow the procedure in Fig.~\ref{fig:blur_decomp}(b) to create a training sample from a series of high-speed video frames.  For instance, as shown for $N=15$ images, we have $\Iblur = (N_1\Ihalf + \Isharp + N_2\Ihalff)/(N_1+N_2+1)$, where $N_1=N_2=7$, $\Iblur = \sum_{n=1}^{15} I[n]/15$, $\Ihalf = \sum_{n=1}^{7} \widehat{I}[n]/7$, $\Ihalff = \sum_{n=9}^{15} \widehat{I}[n]/7$, and $\Isharp = \widehat{I}[8]$. 
We vary $N$ based on the variance of pixel-wise intensities of the chosen video frames along the temporal dimension to aggregate different amounts of blur. The higher the variance, the larger is the motion. We ignore static examples in the training set. Typically the value of $N$ ranged from 11 to 39 frames. We demonstrate our blur decomposition idea further in a discrete time representation in Fig.~\ref{fig:blur_decomp}(b).
We also augment data by randomly employing the following operations: (i) horizontal spatial flipping, (ii) $90^\circ$ or $-90^\circ$ spatial rotations, and (iii) temporal flipping by swapping $\Ishort$ and $\Ishortt$, and $\Ihalf$ and $\Ihalff$, during training. 

To emulate proper short exposure images $\Ishort$ and $\Ishortt$, we add scene-dependent noise according the calibrated noise parameters on-the-fly during training based on the noise model described in \cite{hasinoff:2016:burst}. For the gain level used in our experiments, we calibrate the noise parameters of the camera based on an affine model for the noise variance given by $var(n) := i \alpha  + \beta$, where $i$ is the observed mean intensity, and $\alpha$ and $\beta$ are the calibrated noise parameters. We use the machine vision Blackfly BFS-U3-16S2C camera to capture our short and long exposure images. 

Since our technique uses recursive decomposition, the inputs to the network beyond the first level would have denoised short exposure images. 
However, we observed no significant difference in our test results between employing three separate trained networks with two, one, and zero noisy images for the short-exposure input images appropriately at different decomposition levels, and a single trained network with two noisy images. 
Hence, we report our results for the single trained model approach, that takes in as input  noisy short exposure images, and provide comparisons to the three model approach in the supplemental material.

\subsection{Optimization}
We train the neural network by employing the following costs during optimization: (a) supervised cost and sum cost for $\Ihalf$, $\Isharp$, and $\Ihalff$, and (b) gradient, perceptual, and adversarial costs on the sharp image $\Isharp$. 

The supervised cost is defined as the mean square error between the estimated and ground-truth outputs corresponding to $\Ihalf$, $\Isharp$, and $\Ihalff$. The sum cost is defined by the mean square error according to the blur decomposition process.
The gradient cost is based on the isotropic total-variation norm~\cite{rudin:1992:nonlinear} on the sharp image that encourages sharp edges with homogeneous regions. The perceptual cost 
is defined as the mean squared error between the VGG~\cite{simonyan:2015:very,ledig:2017:photo} features at the conv54 layer of the estimated and ground truth sharp images. We also train a two-class discriminator alongside our network following a generative adversarial training procedure~\cite{goodfellow:2014:generative}, which contributes the generator adversarial cost $p_{adv}$ to encourage the sharp image to lie in the natural image distribution. 

The cost function is given as follows:
{{{    \small
\begin{align}
E = &\|\Ihalf - \Ihalfgt\|_2^2  + \|\Isharp - \Isharpgt\|_2^2 \nonumber \\ &+ \|\Ihalff - \Ihalffgt\|_2^2  \nonumber \\
    &+ \lambda_{sum} \| \Iblur - (N_1 \Ihalf + \Isharp + N_2 \Ihalff) / N \|_2^2 \nonumber \\
    &+ \lambda_{perc} \| \text{VGG}(\Isharp) - \text{VGG}(\Isharpgt) \|_2^2 \nonumber \\
    &+ \lambda_{grad} TV_2(\Isharp) + \lambda_{adv} p_{adv}(\Isharp) \label{eq:optim_cost} 
\end{align} }}}
where 
$\lambda_{sum}$, $\lambda_{perc}$, $\lambda_{adv}$, and $\lambda_{grad}$ are weights of the individual costs, 
$TV_2(\Isharp) = \sum_{i,j} \sqrt{ \nabla_x^2 \Isharp(i,j)  +  \nabla_y^2 \Isharp(i,j) }$ 
is the total variation norm. We use $\lambda_{perc} = 3\times10^{-4}$, $\lambda_{adv} = 10^{-4}$, $\lambda_{grad} = 10^{-4}$, and $\lambda_{sum}=10^{-2}$ in our experiments for the image intensity range $[0,1]$. We train for $10^5$ iterations using Adam as our optimizer with initial learning rate as $10^{-4}$ scaling it by $0.1$ every $2.5\times 10^{4}$ iterations. 

\section{Experiments}
We first demonstrate our blur decomposition through a visualization of blur kernels. We then provide quantitative comparisons with existing methods followed by an analysis on blur amount. Finally, we show qualitative results on real data captured by our cameras.

\begin{figure*}
    \begin{center}
       \renewcommand{\tabcolsep}{1pt}
        \footnotesize
    \begin{tabular}{cccccccccc}
        \includegraphics[width=0.08\linewidth]{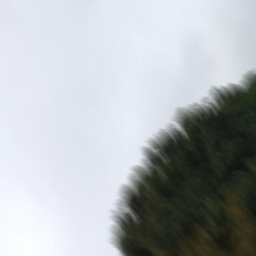} \hspace*{0.15cm}& 
        \includegraphics[width=0.08\linewidth]{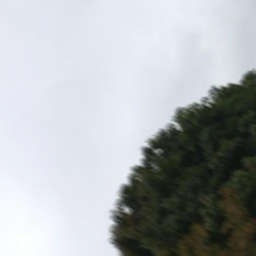}&
        \includegraphics[width=0.08\linewidth]{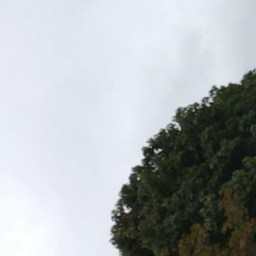} & 
        \includegraphics[width=0.08\linewidth]{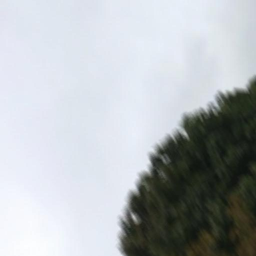} \hspace*{0.15cm}&
        \includegraphics[width=0.08\linewidth]{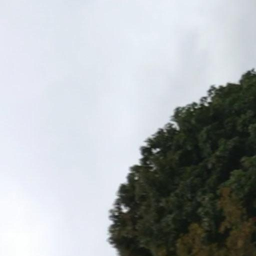}&
        \includegraphics[width=0.08\linewidth]{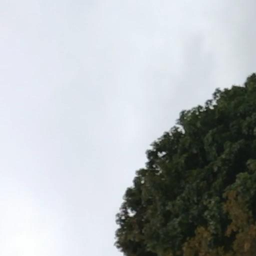} & 
        \includegraphics[width=0.08\linewidth]{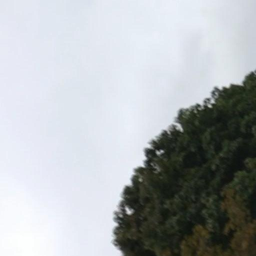} \hspace*{0.15cm}&       
        \includegraphics[width=0.08\linewidth]{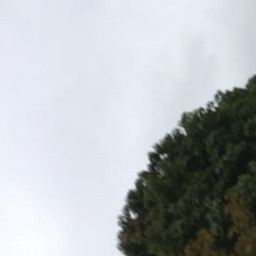}&
        \includegraphics[width=0.08\linewidth]{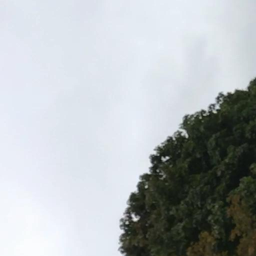} &
        \includegraphics[width=0.08\linewidth]{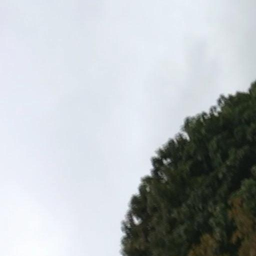}\\
        \includegraphics[width=0.08\linewidth]{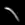} \hspace*{0.15cm}&
        \includegraphics[width=0.08\linewidth]{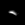} &
        \includegraphics[width=0.08\linewidth]{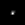}&
        \includegraphics[width=0.08\linewidth]{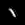} \hspace*{0.15cm}&
        \includegraphics[width=0.08\linewidth]{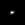}&
        \includegraphics[width=0.08\linewidth]{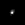} &
        \includegraphics[width=0.08\linewidth]{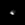} \hspace*{0.15cm}&
        \includegraphics[width=0.08\linewidth]{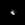}&
        \includegraphics[width=0.08\linewidth]{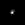}  &
        \includegraphics[width=0.08\linewidth]{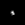}\\
        $\Iblur$ \hspace*{0.15cm} & $\Ihalf$ & $\Isharp$ & $\Ihalff$ \hspace*{0.15cm} 
        &  $\Iquarter$ & $\Isharpx$ & $\Iquarterr$  \hspace*{0.15cm} & $\Iquarterrr$ & $\Isharpy$ & $\Iquarterrrr$\\
        Blurred  \hspace*{0.15cm} & \multicolumn{3}{c}{Level 1}  \hspace*{0.15cm} & \multicolumn{6}{c}{Level 2}
    \end{tabular}
    \end{center}
    \caption{\emph{Successive Reduction of Blur.} At each recursive decomposition level, the blur decreases as depicted in the image patches and blur kernels. The input blurred image has the longest kernel, while it is split into two in the first level and further into four in the second level. The deblurred images at both levels have close-to-delta kernels indicating sharpness of deblurred images.}
    \label{fig:exp:blurkernels}
 \end{figure*}

\subsection{Successive Reduction of Blur} 

We demonstrate our blur decomposition through blur kernels estimated using the state-of-the-art blind deblurring method of \cite{pan:2016:blind}. A blur kernel describes the motion experienced by a point light source located at the image center, and thus acts as an indicator of residual blur/motion present in the image.
Fig.~\ref{fig:exp:blurkernels}(a) shows a patch from a motion blurred image $\Iblur$ 
and its long blur kernel. 
The network takes this image and the short exposure images as inputs (not shown). At first level of decomposition, 
the half-blurred images have blur kernels of reduced length indicating shorter blur. One can neatly trace down the long kernel trajectory of $\Iblur$ by conjoining the split trajectories of level-1. 
(The kernel estimation is blind, and therefore, it is always centered.) Also, the blur kernel of the middle image $\Isharp$ has a close-to-delta kernel indicating a negligible blur.  
Similarly, in the second level, we get four half-blurred images with further reduction of blur and the corresponding two deblurred images. Thus, our recursive decomposition provides an elegant way to remove blur and achieve our goal. 

\begin{figure*}
    \begin{center}
        \renewcommand{\tabcolsep}{1pt}
        \footnotesize
        \begin{tabular}{ccccccc}
            \includegraphics[width=0.14\linewidth]{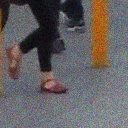}&
            \includegraphics[width=0.14\linewidth]{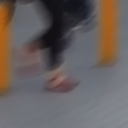}&
            \includegraphics[width=0.14\linewidth]{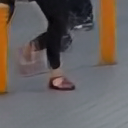}&
            \includegraphics[width=0.14\linewidth]{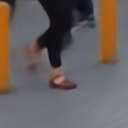}&
            \includegraphics[width=0.14\linewidth]{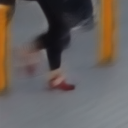}&
            \includegraphics[width=0.14\linewidth]{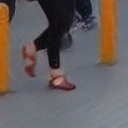}&
            \includegraphics[width=0.14\linewidth]{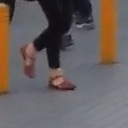}\\
            \includegraphics[width=0.14\linewidth]{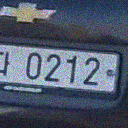}&
            \includegraphics[width=0.14\linewidth]{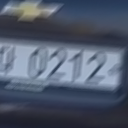}&
            \includegraphics[width=0.14\linewidth]{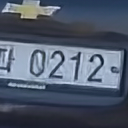}&            
            \includegraphics[width=0.14\linewidth]{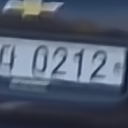}&
            \includegraphics[width=0.14\linewidth]{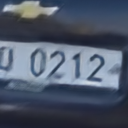}&
            \includegraphics[width=0.14\linewidth]{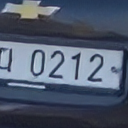}&
            \includegraphics[width=0.14\linewidth]{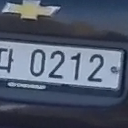}\\
            Short/Noisy & Long/Blurry & Single-blur\cite{jin:2018:learning} & Multi-blur\cite{jin:2019:learning} &
            Ours (L-only) & Ours (S-L-S) & Ground-truth
        \end{tabular}
    \end{center}
    \caption{\textit{Quasi-real blur and noise on high-speed video GoProTest \cite{nah:2017:deep} data.} Comparisons with single-blur sequencing and multi-blur sequencing prior works.}
    \label{fig:gopro}
\end{figure*}

\begin{figure}
    \begin{center}
        \renewcommand{\tabcolsep}{0.01pt}
        \footnotesize
        \begin{tabular}{cccccc}
            \multicolumn{2}{c}{1ms-15ms} & 
            \multicolumn{2}{c}{1ms-25ms} & 
            \multicolumn{2}{c}{1ms-32ms}\\
            \includegraphics[width=0.166\linewidth]{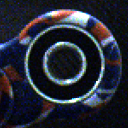}&
            \includegraphics[width=0.166\linewidth]{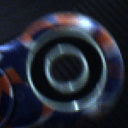}\hspace*{0.1cm}&
            \includegraphics[width=0.166\linewidth]{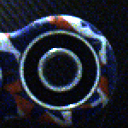}&
            \includegraphics[width=0.166\linewidth]{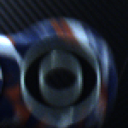}\hspace*{0.1cm}&
            \includegraphics[width=0.166\linewidth]{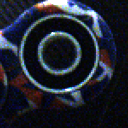}&
            \includegraphics[width=0.166\linewidth]{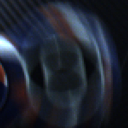}
        \end{tabular}
        \begin{tabular}{cccccc}
            \multicolumn{6}{c}{(a) Short and Long exposure images for different exposure intervals.}\\
            \includegraphics[width=0.166\linewidth]{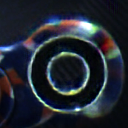}&
            \includegraphics[width=0.166\linewidth]{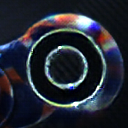}\hspace*{0.1cm}&
            \includegraphics[width=0.166\linewidth]{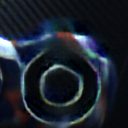}&
            \includegraphics[width=0.166\linewidth]{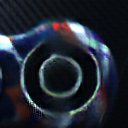}\hspace*{0.1cm}&
            \includegraphics[width=0.166\linewidth]{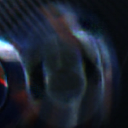}&
            \includegraphics[width=0.166\linewidth]{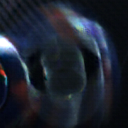}\\        
            \multicolumn{6}{c}{(b) Single-blur sequencing outputs \cite{jin:2018:learning} (takes a single blurred image as input)}\\
            \includegraphics[width=0.166\linewidth]{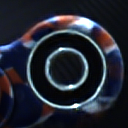}&
            \includegraphics[width=0.166\linewidth]{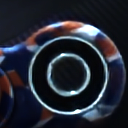}\hspace*{0.1cm}&
            \includegraphics[width=0.166\linewidth]{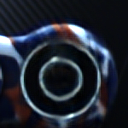}&
            \includegraphics[width=0.166\linewidth]{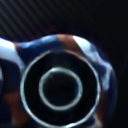}\hspace*{0.1cm}&
            \includegraphics[width=0.166\linewidth]{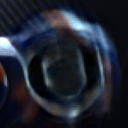}&
            \includegraphics[width=0.166\linewidth]{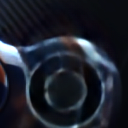}\\        
            \multicolumn{6}{c}{(c) Multi-blur sequencing outputs \cite{jin:2019:learning} (takes four blurred images as inputs)}\\
            \includegraphics[width=0.166\linewidth]{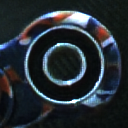}&
            \includegraphics[width=0.166\linewidth]{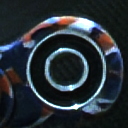}\hspace*{0.1cm}&
            \includegraphics[width=0.166\linewidth]{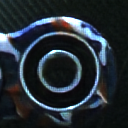}&
            \includegraphics[width=0.166\linewidth]{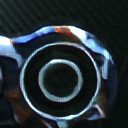}\hspace*{0.1cm}&
            \includegraphics[width=0.166\linewidth]{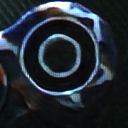}&
            \includegraphics[width=0.166\linewidth]{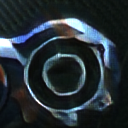}\\
            \multicolumn{6}{c}{(d) Proposed short-long-short sequencing outputs}
        \end{tabular}        
    \end{center}
    \caption{\textit{Real Spinner data.} Comparisons with single and multi-blur photosequencing prior works for different blur amounts.}
    \label{fig:spinner}
\end{figure}

\begin{figure}
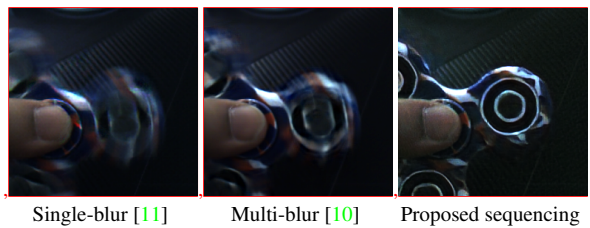

    \footnotesize
    \begin{center}
        \renewcommand{\tabcolsep}{0.1pt}
        \begin{tabular}{ccc}
            \animbox{\animategraphics[width=0.3\linewidth,loop]{5}{img/spinner/spinner7_gamma/crop3/Sequencer-18575809-3-esti}{1}{4}}&
            \animbox{\animategraphics[width=0.3\linewidth,loop]{5}{img/spinner/spinner7_gamma/crop3/}{0000}{0004}}&
            \animbox{\animategraphics[width=0.3\linewidth,loop]{5}{img/spinner/spinner7_gamma/crop3/anim/3_c}{001}{005}}\\
            Single-blur \cite{jin:2018:learning} & Multi-blur \cite{jin:2019:learning} & Proposed sequencing
        \end{tabular}
    \end{center}
    \caption{Temporal ordering in photosequencing. Click to play.}
    \label{fig:spinner_order}
\end{figure}

\begin{table}
    \caption{PSNRs (dB) on GoProTest~\cite{nah:2017:deep} for blur  of 11 frames.}
   \small
    \begin{center}
        \begin{tabular}{@{}cccc@{}}
            \hline 
            Method & $I_{0.25}$ & $I_{0.5}$ & $I_{0.75}$ \\
            \hline 
            Single-blur sequencing\cite{jin:2018:learning} & 27.24 & 28.81 & 27.83\\ 
            Multi-blur sequencing\cite{jin:2019:learning} & 29.38  & 29.52 & 29.41\\
            Ours (long only)& 27.13 & 27.81 & 27.32 \\
            Ours (short-long-short)& \textbf{30.38} & \textbf{30.77} & \textbf{30.22}\\
            \hline
        \end{tabular}
    \end{center}
    \label{tab:quant}
\end{table}

\begin{table}
    \caption{PSNR (dB) of $\Isharp$ based on blur frame length.}
    \small
    \begin{center}
        \begin{tabular}{@{}ccccc@{}}
            \hline 
            Blur frame length & 9 &  11 & 15 & 19\\
            \hline 
            Multi-blur sequencing\cite{jin:2019:learning} & 30.31 & 29.52 & 29.34 & 29.13\\
            Ours (short-long-short)& \textbf{30.87}& \textbf{30.77} & \textbf{30.62} & \textbf{30.21}\\
            \hline
        \end{tabular}
    \end{center}
    \label{tab:quant2}
\end{table}

\begin{figure*}
    \begin{center}
        \renewcommand{\tabcolsep}{1pt}
        \footnotesize
        \begin{tabular}{cccccccc}
        \includegraphics[width=0.125\linewidth]{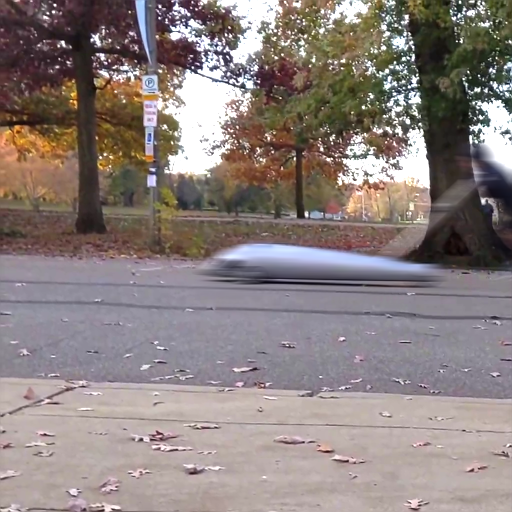}&
        \includegraphics[width=0.125\linewidth]{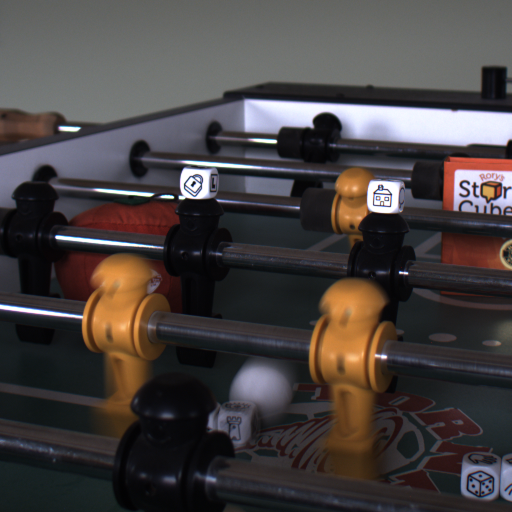}&
        \includegraphics[width=0.125\linewidth]{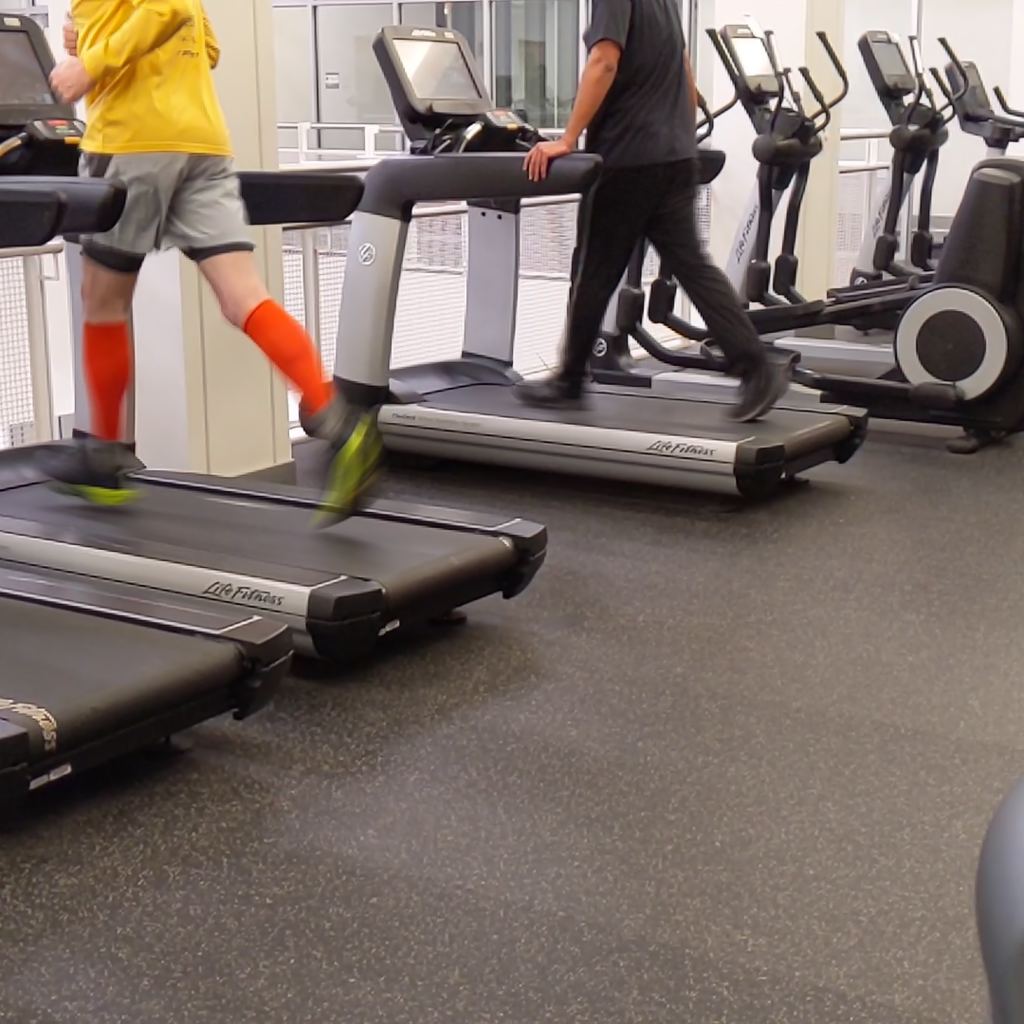}&
        \includegraphics[width=0.125\linewidth]{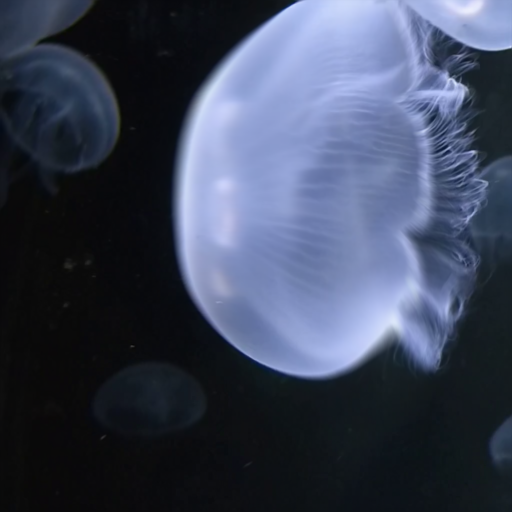}&
        \includegraphics[width=0.125\linewidth]{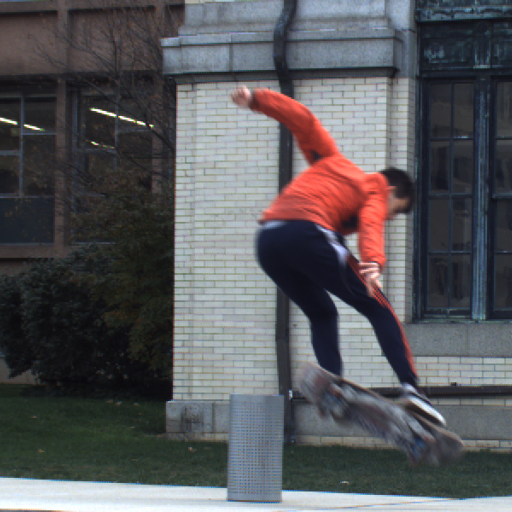}&
        \includegraphics[width=0.125\linewidth]{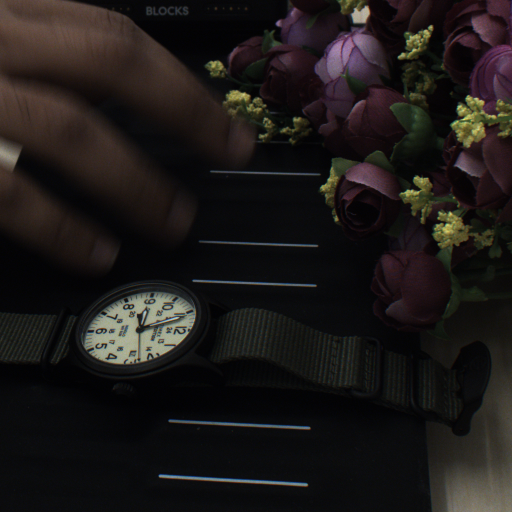}&
        \includegraphics[width=0.125\linewidth]{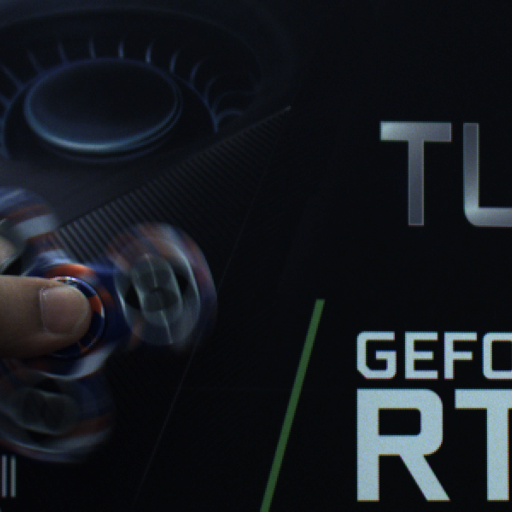}&
        \includegraphics[width=0.125\linewidth]{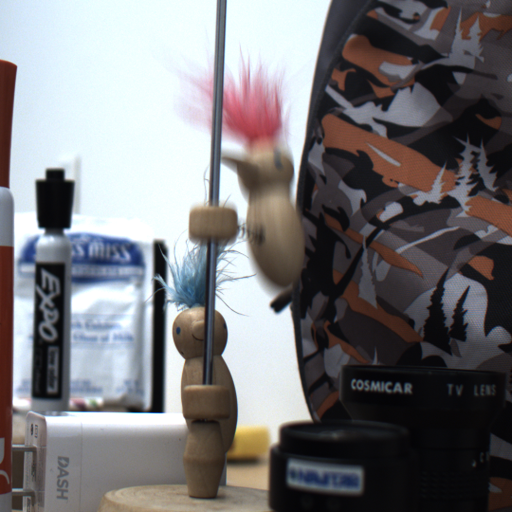}\\
        Race & Foosball & Gym & Jellyfish & Skate & Keyboard & Spinner & Pecker
        \end{tabular}
    \end{center}
    \caption{\textit{Some scenes from our real data captures.} The whole sequence is available as supplementary.}
    \label{fig:scene_examples}
\end{figure*}

\begin{figure*}
    \begin{center}
    \renewcommand{\tabcolsep}{1pt}
    \footnotesize
    \hspace*{1cm}\includegraphics[width=0.9\linewidth]{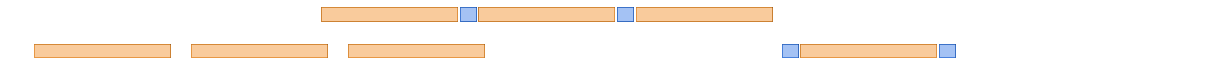}\\
    \begin{tabular}{cccccc}
    Long & Long & Long  &Short & Long & Short \\
    \includegraphics[width=0.15\linewidth]{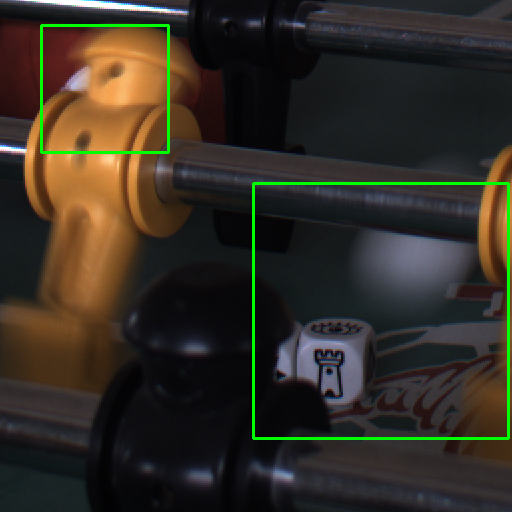}&
    \includegraphics[width=0.15\linewidth]{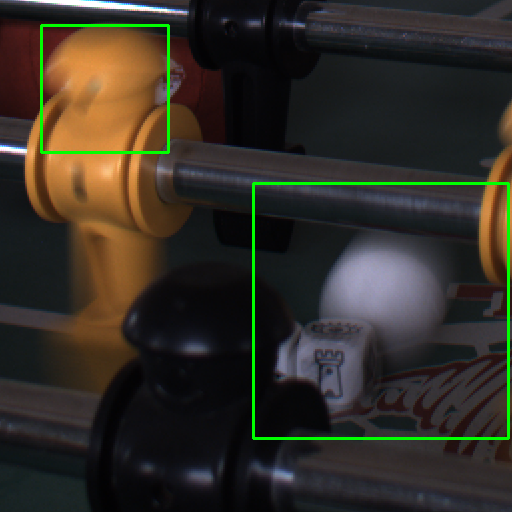}&
    \includegraphics[width=0.15\linewidth]{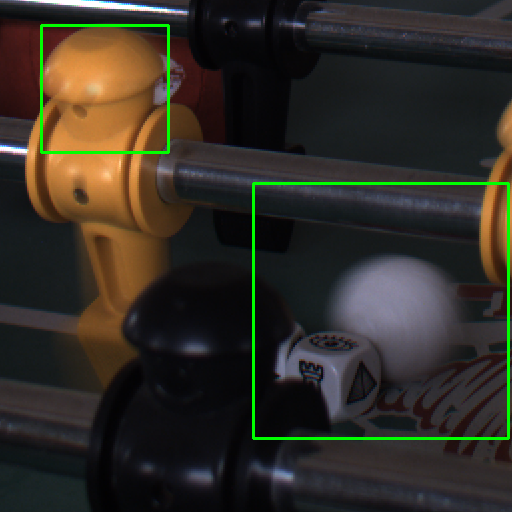}\ \ &
    \includegraphics[width=0.15\linewidth]{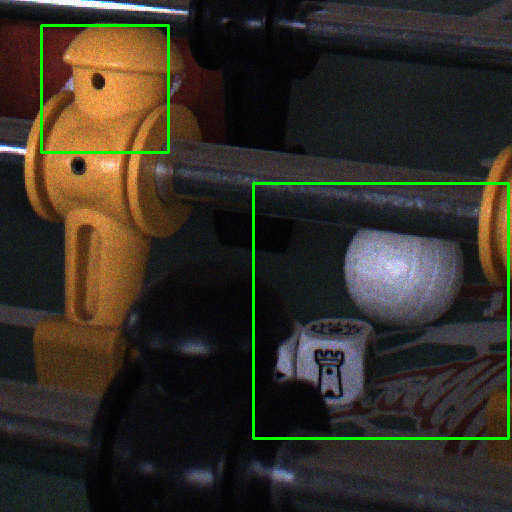}&
    \includegraphics[width=0.15\linewidth]{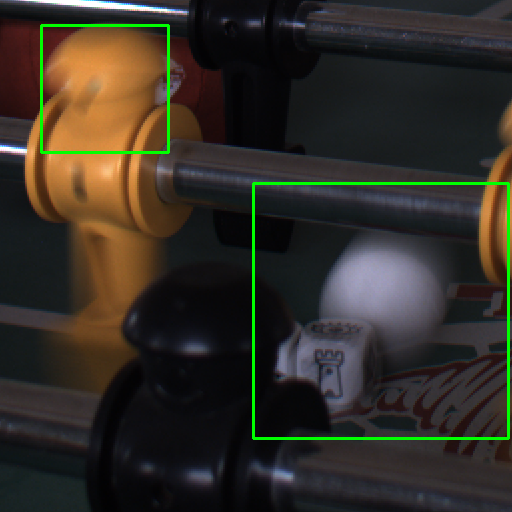}&
    \includegraphics[width=0.15\linewidth]{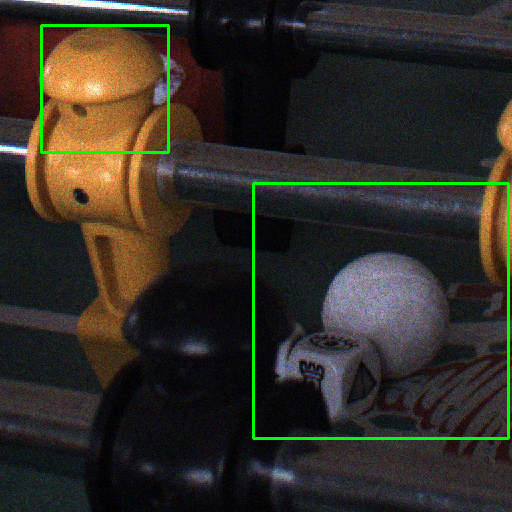}
    \end{tabular}
    \\
    \begin{tabular}{cccccccccccc}
        \includegraphics[width=0.065\linewidth]{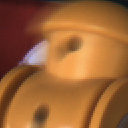}&
        \includegraphics[width=0.065\linewidth]{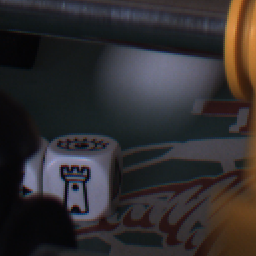}&
        \includegraphics[width=0.065\linewidth]{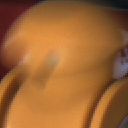}&
    \includegraphics[width=0.065\linewidth]{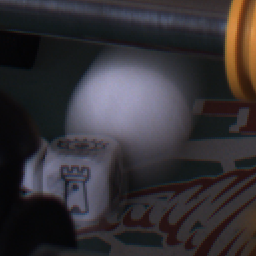}&
    \includegraphics[width=0.065\linewidth]{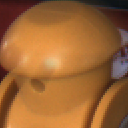}&
    \includegraphics[width=0.065\linewidth]{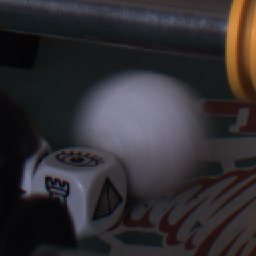} \hspace*{0.8cm} &
    \includegraphics[width=0.065\linewidth]{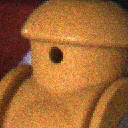}&
    \includegraphics[width=0.065\linewidth]{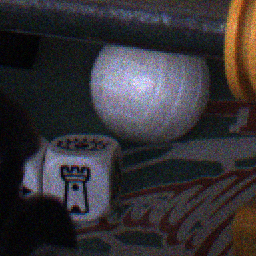}&
    \includegraphics[width=0.065\linewidth]{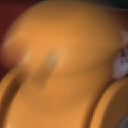}&
    \includegraphics[width=0.065\linewidth]{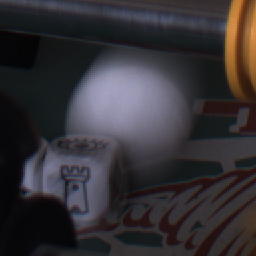}&
    \includegraphics[width=0.065\linewidth]{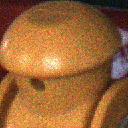} &
    \includegraphics[width=0.065\linewidth]{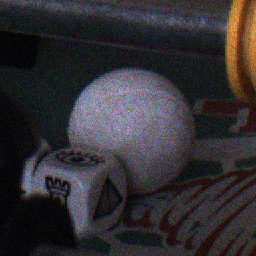}       
    \end{tabular}
    \\ 
    \includegraphics[width=0.065\linewidth]{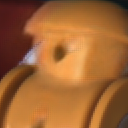}
    \includegraphics[width=0.065\linewidth]{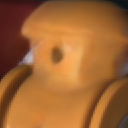}
    \includegraphics[width=0.065\linewidth]{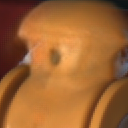}
    \includegraphics[width=0.065\linewidth]{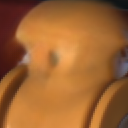}
    \includegraphics[width=0.065\linewidth]{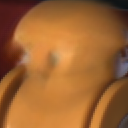}
    \includegraphics[width=0.065\linewidth]{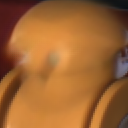}
    \includegraphics[width=0.065\linewidth]{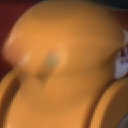} \ \ 
    \includegraphics[width=0.065\linewidth]{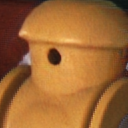}
    \includegraphics[width=0.065\linewidth]{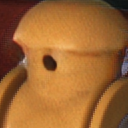}
    \includegraphics[width=0.065\linewidth]{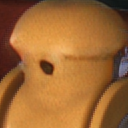}
    \includegraphics[width=0.065\linewidth]{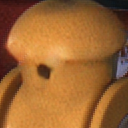}
    \includegraphics[width=0.065\linewidth]{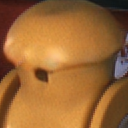}
    \includegraphics[width=0.065\linewidth]{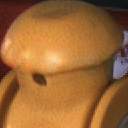}
    \includegraphics[width=0.065\linewidth]{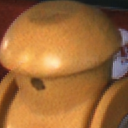}        
    \\
    \includegraphics[width=0.065\linewidth]{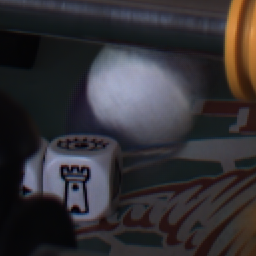}
    \includegraphics[width=0.065\linewidth]{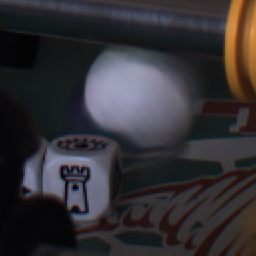}
    \includegraphics[width=0.065\linewidth]{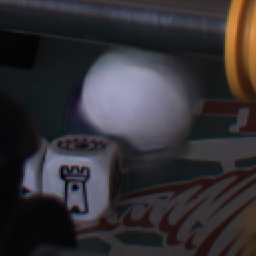}
    \includegraphics[width=0.065\linewidth]{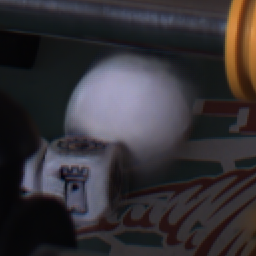}
    \includegraphics[width=0.065\linewidth]{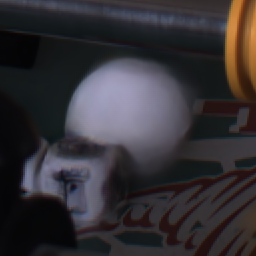}
    \includegraphics[width=0.065\linewidth]{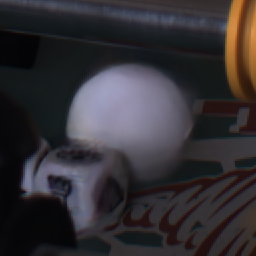}
    \includegraphics[width=0.065\linewidth]{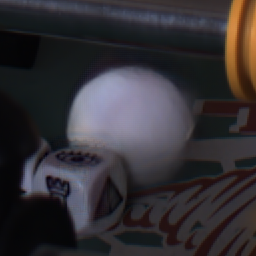}  \ \ 
    \includegraphics[width=0.065\linewidth]{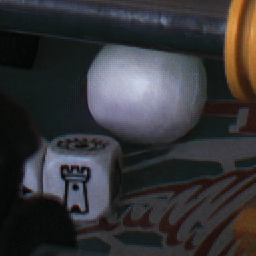}
    \includegraphics[width=0.065\linewidth]{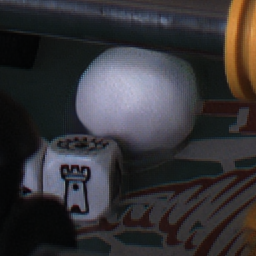}
    \includegraphics[width=0.065\linewidth]{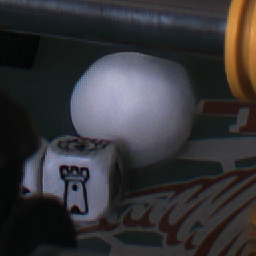}
    \includegraphics[width=0.065\linewidth]{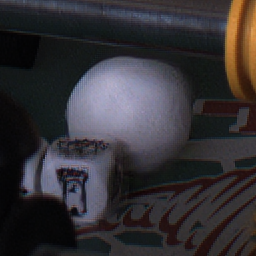}
    \includegraphics[width=0.065\linewidth]{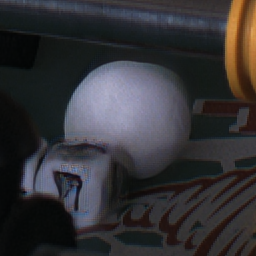}
    \includegraphics[width=0.065\linewidth]{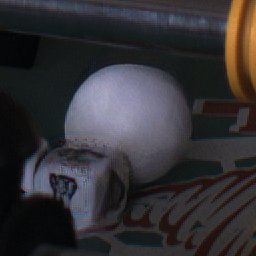}
    \includegraphics[width=0.065\linewidth]{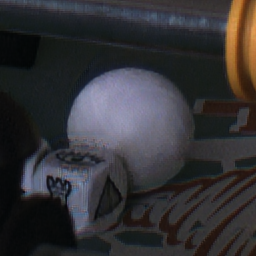}    
    \\
    \begin{tabular}{cc}
        (a) Multi-blur sequencing \cite{jin:2019:learning} \hspace*{3cm} & (b) Proposed short-long-short sequencing 
    \end{tabular}
    \end{center}
    \caption{\textit{Comparison with multi-blur sequencing technique on real Foosball data.} Recovering the sequence from multiple blurred images~\cite{jin:2019:learning} lose out on sharp textures albeit correct motion sequencing. Our short-long-short exposure inputs recover both the motion and texture successfully.}
    \label{fig:foosball}
 \end{figure*}

\subsection{Quantitative Analysis} 
We
analyze the performance of 3x sequence-rate level, i.e. the PSNR of the middle deblurred image $\Isharp$ and that of the second-level decomposition images, $\Isharpx$ and $\Isharpy$ against 
the single-blur sequencing of \cite{jin:2018:learning} and multi-blur sequencing of \cite{jin:2019:learning}. 
In addition, we also consider an ablation case of our method by considering only the long exposure blurred image as input without any short exposure images, indicated by \textit{long-only}.

Since each of these methods need different types of inputs, we prepare the testing set as follows. We created a set of eleven examples from the eleven test videos of the GoPro 240fps data~\cite{nah:2017:deep}. Each example is a sequence of alternating short and long exposure images. The short exposure is created by taking a single video frame, while the long exposure is synthesized as an average of 11 frames. Our method takes a single consecutive short-long-short triplet as input with added noise to the short exposures. The single-blur sequencing method takes only a single long exposure image, while multi-blur sequencing takes four long exposure images as inputs. 

The results of our analysis are provided in Table \ref{tab:quant}. First, our network trained with short-long-short exposure inputs performs better than training with only the long exposure image indicating the benefit of capturing additional short exposure images in photosequencing. 
The multi-blur sequencing performs better than the single-blur sequencing owing to more available information as expected. In turn, we are able to perform better than the multi-blur sequencing. Our method recovers textures missed in heavy blur from the short exposure images. Fig.~\ref{fig:gopro}(top) depicts this behavior where the prior works are able to reconstruct the leg on the ground which has lesser blur, while the other leg is not recovered even by the multi-blur technique. Our output shows better textures with minimal residual blur and is practically noise-free.

\subsection{Blur Amount Analysis}
The amount of blur observed in the long exposure image is a synergy of exposure time and motion. We repeated our experiments on the test data for different frame lengths shown in Table \ref{tab:quant2}. The multi-blur sequencing method performed better for shorter frame lengths almost on par with our method and worse for the longer ones.  

Fig.~\ref{fig:spinner}(a) shows our real captures of a spinner for multiple exposure configurations. At the lowest long exposure, the blur is minimal, while at the longest setting, the spinner texture is almost lost. Three frames from sequencing results of different methods are shown in (b,c) and (d). Single-blur technique (b) shows residual blur artifacts even at medium blur shown in the third column-set. While the multi-blur technique (c) could recover some texture at the highest setting, our method outperforms it. Further, single-blur technique could not disambiguate temporal ordering leading to wrong local spinner motion as shown in Fig.~\ref{fig:spinner_order}, while our short-long-short sequencing correctly recovers the ordering.

 \begin{figure*}
    \begin{center}
        \renewcommand{\tabcolsep}{1pt}
        \footnotesize
        \begin{tabular}{cccccc}
        \includegraphics[width=0.165\linewidth]{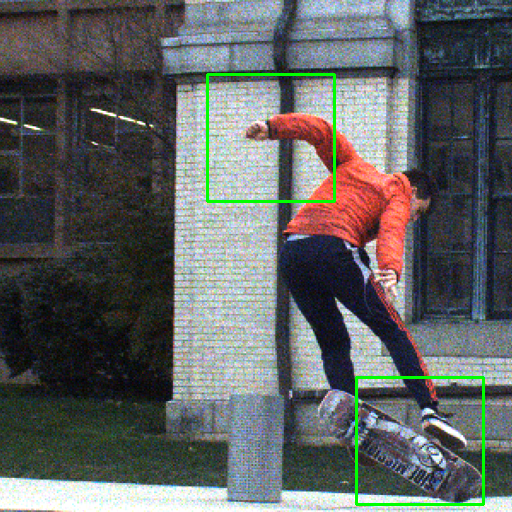}&
        \includegraphics[width=0.165\linewidth]{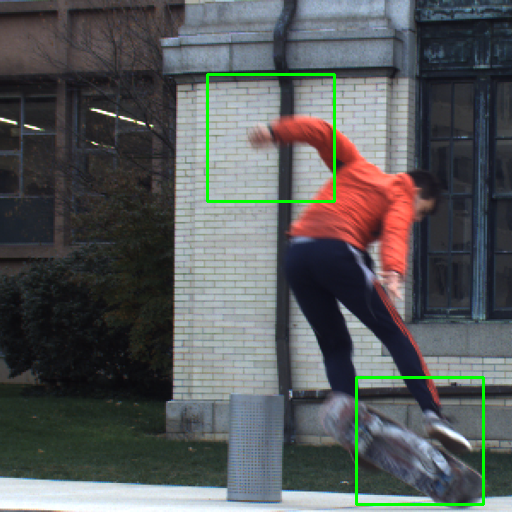}&
        \includegraphics[width=0.165\linewidth]{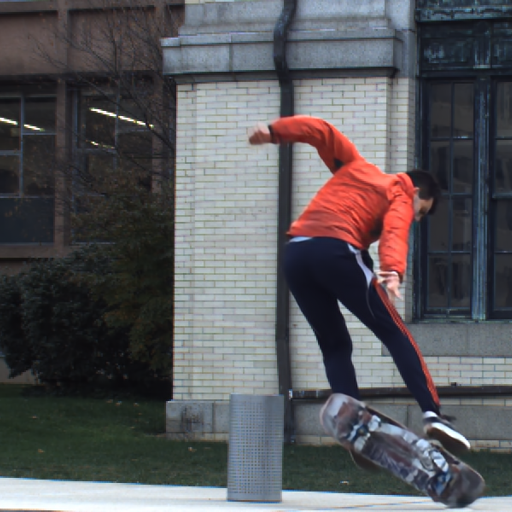}&
        \includegraphics[width=0.165\linewidth]{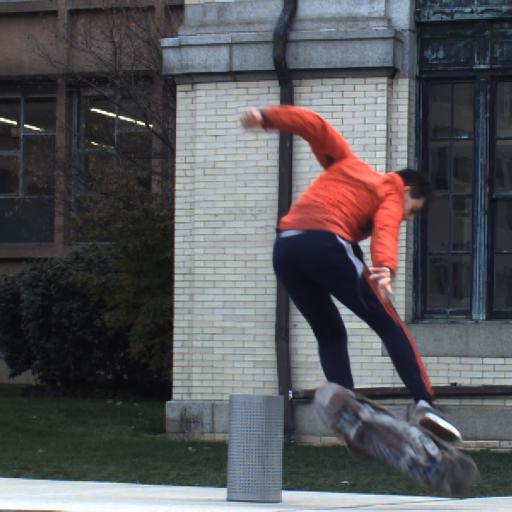}&
        \includegraphics[width=0.165\linewidth]{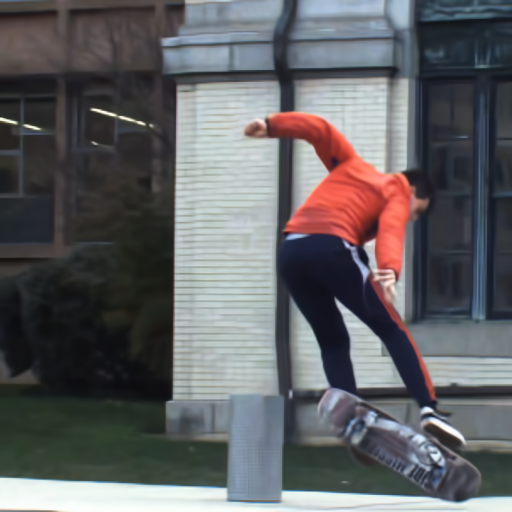}&        
        \includegraphics[width=0.165\linewidth]{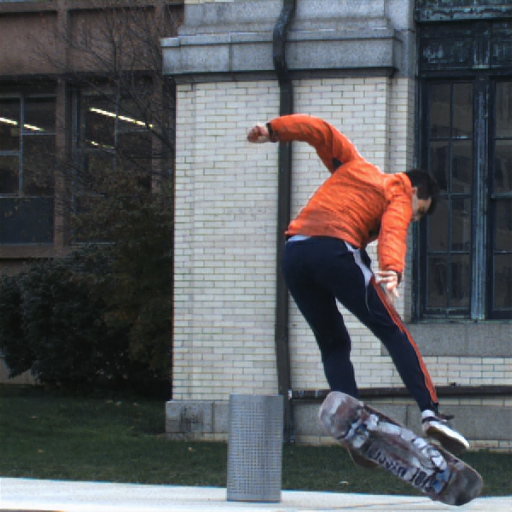}\\
        \includegraphics[width=0.165\linewidth]{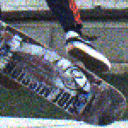}&
        \includegraphics[width=0.165\linewidth]{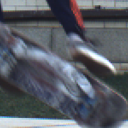}&
        \includegraphics[width=0.165\linewidth]{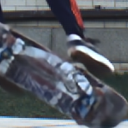}&
        \includegraphics[width=0.165\linewidth]{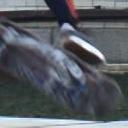}&
        \includegraphics[width=0.165\linewidth]{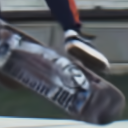}&        
        \includegraphics[width=0.165\linewidth]{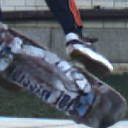}\\
        \includegraphics[width=0.165\linewidth]{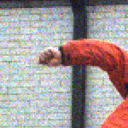}&
        \includegraphics[width=0.165\linewidth]{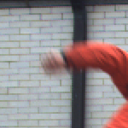}&
        \includegraphics[width=0.165\linewidth]{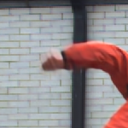}&
        \includegraphics[width=0.165\linewidth]{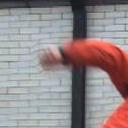}&
        \includegraphics[width=0.165\linewidth]{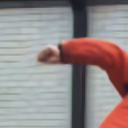}&
        \includegraphics[width=0.165\linewidth]{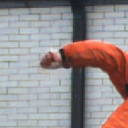}\\
        (a) Short & (b) Long & (c) Multi-blur sequencing \cite{jin:2019:learning} & (d) Deblur+Interp \cite{tao:2018:scale,jiang:2018:super} & (e) Denoise+Interp \cite{dabov:2007:image,jiang:2018:super} & (f) Ours
        \end{tabular}
    \end{center}
    \caption{\textit{Results on Skate data.} Our method correctly transfers textures of static regions (wall linings) from the blurred image and textures of moving objects (hand-fist) from the short exposure image.}
    \label{fig:skate}
\end{figure*}

\begin{figure}
    \begin{center}
        \renewcommand{\tabcolsep}{0.5pt}
        \footnotesize
        \begin{tabular}{cc}
            \includegraphics[width=0.45\linewidth]{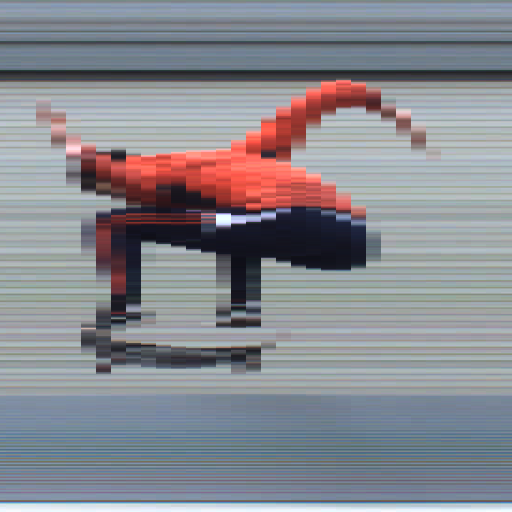}&
            \includegraphics[width=0.45\linewidth]{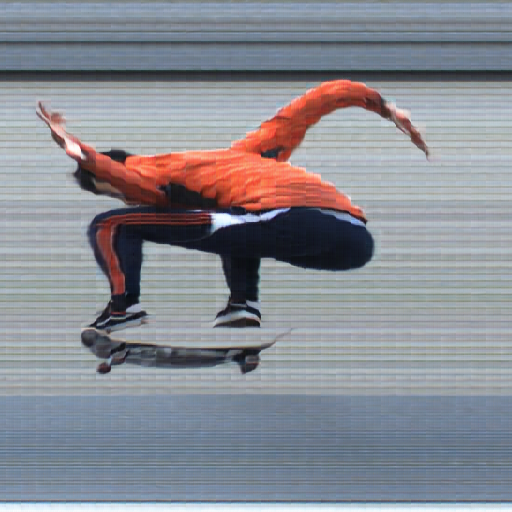}\\
            \includegraphics[width=0.45\linewidth]{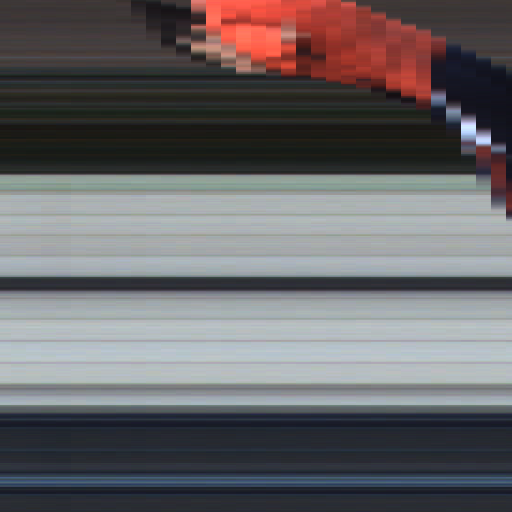}&
            \includegraphics[width=0.45\linewidth]{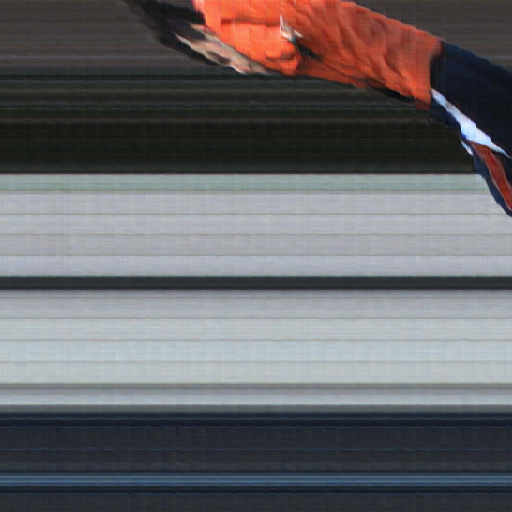}\\
            (a) Input capture & (b) Recovered photosequence
        \end{tabular}
    \end{center}
    \caption{\textit{Time slices on Skate.} We recovered the photosequence of consecutive short and long exposures using our method. XT (top) and YT (bottom) slices show discontinuites in the horizontal time axis for the input capture indicating lower frame rate. The slices of our photosequencing result on the right show smooth interpolation of motion during the long exposure intervals.}
    \label{fig:skate_slice}
\end{figure}

\begin{figure}
    \begin{center}
        \renewcommand{\tabcolsep}{1pt}
        \footnotesize
        \begin{tabular}{cccc}
        \includegraphics[width=0.24\linewidth]{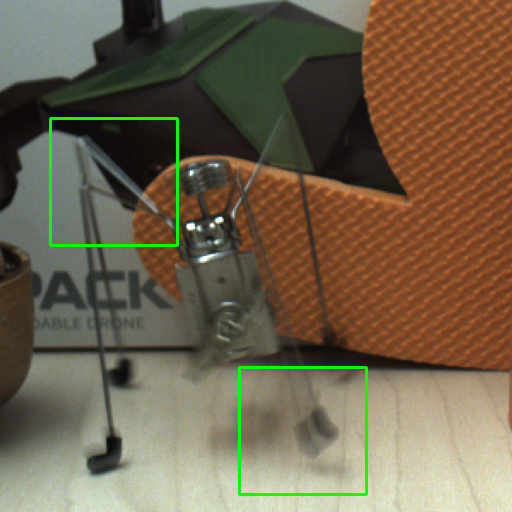}&
        \includegraphics[width=0.24\linewidth]{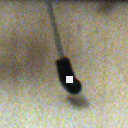}&
        \includegraphics[width=0.24\linewidth]{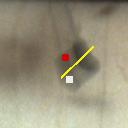}&        
        \includegraphics[width=0.24\linewidth]{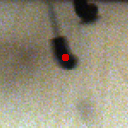}\\
        (a) Blurred & \multicolumn{3}{c}{(b) Short-long-short crops}
        \end{tabular}       
        \begin{tabular}{cccccc}
        \includegraphics[width=0.15\linewidth]{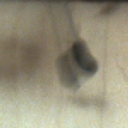}&
        \includegraphics[width=0.15\linewidth]{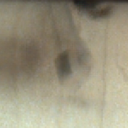}&
        \includegraphics[width=0.15\linewidth]{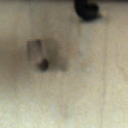}\hspace*{0.15cm}&        
        \includegraphics[width=0.15\linewidth]{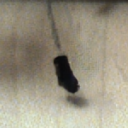}&
        \includegraphics[width=0.15\linewidth]{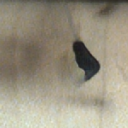}&
        \includegraphics[width=0.15\linewidth]{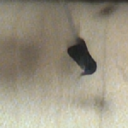}\\
        \multicolumn{3}{c}{(b) Multi-blur sequencing~\cite{jin:2019:learning}} & \multicolumn{3}{c}{(c) Proposed  sequencing}
    \end{tabular}
\end{center}
\caption{\textit{Results on JumpingToy data.} Very fast jumping of the thin-legged object misaligns the line of blur with the short exposures as in (b). This causes imperfect reconstructions by our method, and so does multi-blur sequencing as shown in (b,c).}
\label{fig:jumpingtoy}
\end{figure}

\subsection{Results on Real Data}

We capture sequences of short and long exposures of a wide variety of scenes using our Blackfly camera to show the efficacy of our photosequencing, some of which are shown in Fig.~\ref{fig:scene_examples}. Our captures comprise both indoor and outdoor scenes with different types of motions including linear blur in \textit{Race}, rotations in \textit{Foosball}, human nonlinear motions in \textit{Gym} and \textit{Skate}, fine textured motion of \textit{Jellyfish}, and closeup finger movement in \textit{Keyboard}. Note that we cannot use any of the existing real blurred videos since our technique requires short exposure images as well. 

\textit{Comparison with multi-blur sequencing.} 
Fig.~\ref{fig:foosball} shows the \textit{Foosball} scene with noisy short exposures, and blurry player and ball in long exposures. (Please zoom-in to see noise and blur clearly in first two rows.) The third row and fourth row show the cropped results of \cite{jin:2019:learning} and our method. \cite{jin:2019:learning} recovers the direction of motion correctly but the blur is not completely removed. Our method correctly recovers the sharpness of the player and the ball as well as the motion ordering. 

\textit{Comparison with frame interpolation.} 
Fig.~\ref{fig:skate} shows input images and crops of the Skate scene in (a,b). 
in which heavy noise and blur can be observed. 
The multi-blur sequencing fails to recover intricate textures of the skateboard and fist as seen in (c). 
We follow two pipelines to compare with frame interpolation \cite{jiang:2018:super}. First, we deblur two long exposures using the state-of-the-art deblurring method of \cite{tao:2018:scale} and interpolate frames. Second, we denoise two short exposures using 
BM3D \cite{dabov:2007:image}
followed by frame interpolation. In the first case result shown in (d), the heavy residual blur due to poor deblurring of heavy motion is passed on to interpolation which can only recover blurred frames. In the second case (e), we do note that the texture in the fist is devoid of blur since it uses only the short exposure frames, however the denoised image has a washed-out appearance. Further, the static wall undergoes noise-denoise process for no reason only to have its texture erased. Our method correctly exploits textures of static regions (wall) from the blurred image and textures of moving objects (hand, fist) from the short exposure image as shown in (f).

The proposed technique recovers a sharp frame sequence for a short-long-short triplet. By combining the outputs from a longer sequence of alternating short and long exposures, we can produce high frame-rate videos of long time durations.
We showcase some examples of this in the supplemental video. We do note that such sequences have temporal tiling artifacts since each triplet leads to sequences in isolation. We show XT and YT slices for the result of \textit{Skate} data over multiple triplets in Fig.~\ref{fig:skate_slice}.

\textit{Failure Example.} A challenging scenario is that of heavy motion of thin structures. Fig.~\ref{fig:jumpingtoy}(a) portrays the \textit{JumpingToy} scene wherein the thin legs of the jumping metal toy move very fast. The motion is so ragged that the line of blur marked in yellow is disconnected from the two short exposure points (white to red) as shown in the image crops of Fig.~\ref{fig:jumpingtoy}(a). Our method shows artifacts in the photosequence with partial leg reconstructions as shown in (c-top). In comparison, the multi-blur sequencing method performs worse as shown in (b), and suffers residual blur artifacts in other regions as well where we perform better as shown in the last row.

\section{Conclusion}
We proposed a new technique to record fast motion events by capturing short-long-short exposure images. 
We utilize their complementary information of texture and motion in these images to estimate the sharp high frame-rate photosequence associated with the long exposure. 
Our technique provides an approach that can be easily implemented on  mobile devices, thereby providing the capability of high-speed capture with little change to existing hardware.
Hence, we believe this would be a fascinating new application of mobile computational photography.

\paragraph{Acknowledgement}
The work of VR, SZ, and ACS was supported in part by the ARO Grant W911NF-15-1-0126, a gift from Samsung, and the Intelligence Advanced Research Projects Activity (IARPA) via Department of Interior/Interior Business Center (DOI/IBC) contract number D17PC00345. The U.S.~Government is authorized to reproduce and distribute reprints for Governmental purposes notwithstanding any copyright annotation thereon. Disclaimer: The views and conclusions contained herein are those of the authors and should not be interpreted as necessarily representing the official policies or endorsements, either expressed or implied of IARPA, DOI/IBC or the U.S.~Government.

{\small
\bibliographystyle{ieee_fullname}
\bibliography{ms}
}

\definecolor{downstride2}{rgb}{0.6,0.1,0.1}
\definecolor{upstride2}{rgb}{0.1,0.6,0.1}

\renewcommand\thefigure{S\arabic{figure}} 
\renewcommand\thetable{S\arabic{table}} 
\renewcommand\thesection{S\arabic{section}} 
\makeatother

\newpage

\title{Photosequencing of Motion Blur using Short and Long Exposures\\Supplementary Material}
\author{}
\date{}
\maketitle
\ \\[-1.5cm]
In this supplementary PDF, we provide the details of our neural network architecture, and comparison between inference approaches based on single and three trained models. 

\section{Comparison of Single and Three Model Approaches}
\begin{figure}
    \begin{center}
        \footnotesize
        \renewcommand{\tabcolsep}{0.5pt}
        \begin{tabular}{c}
            \includegraphics[width=\linewidth]{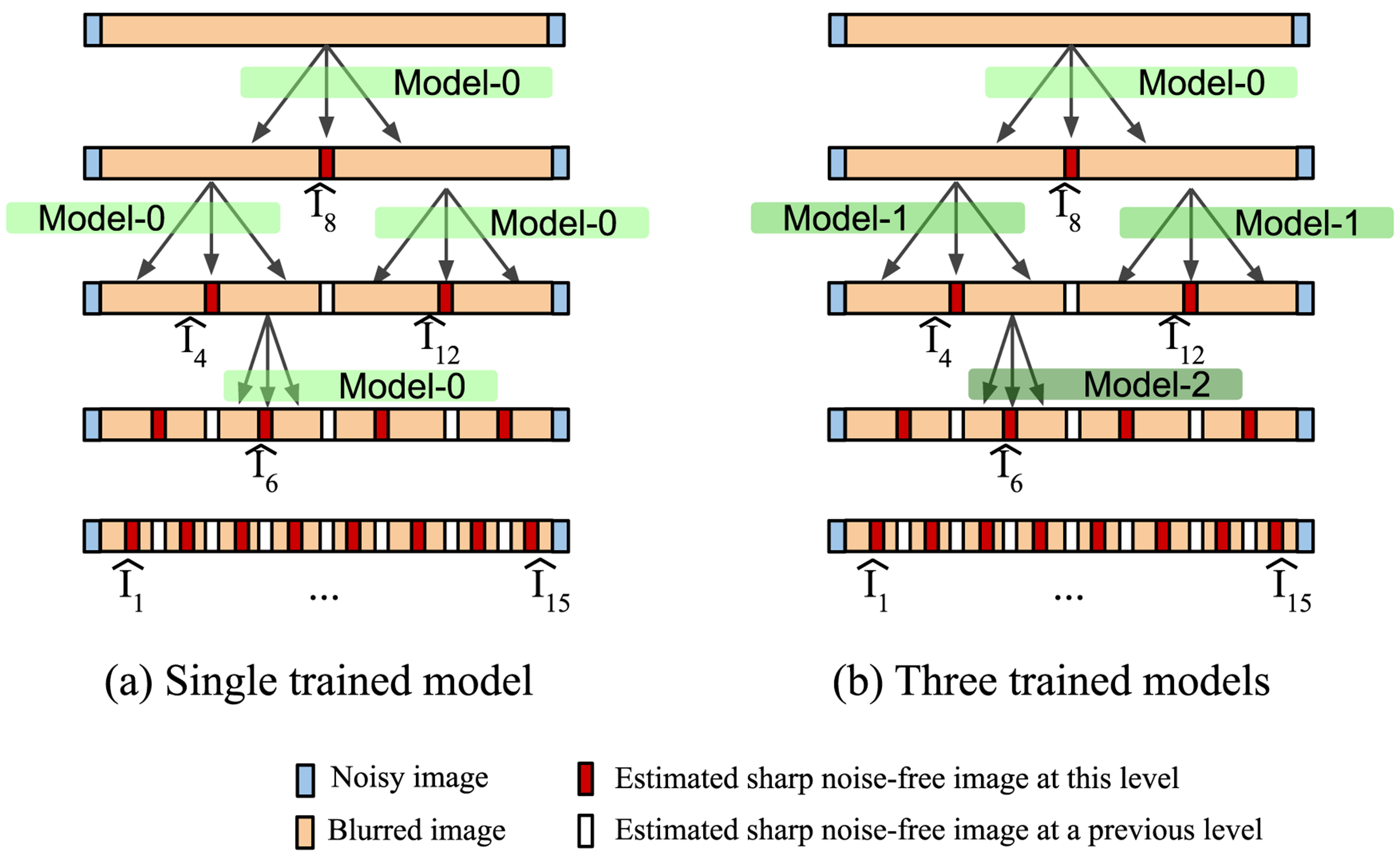}
        \end{tabular}
    \end{center}
    \caption{\textit{Single and three trained model approaches.} (a) In the first approach, a single model is trained with two short exposure noisy images, and the same model is used at inference for all levels of decomposition even if one or both of the short exposure inputs are noise-free estimated images from a previous level. (b) In the second approach three models are trained with two, one, and zero noisy images as inputs. The respective model is used based on the number of estimated noise-free images involved for that particular inference set of inputs. For instance, the estimations of $I_4$ and $I_{12}$ in (b) use Model-1 since it involves the estimated noise-free $I_8$ as one of its inputs and the other short-exposure input is noisy, and the estimation of $I_6$ uses Model-2 since both its short-exposure inputs $I_4$ and $I_8$ are noise-free.}
    \label{fig:mult_models}
\end{figure}

As mentioned in Sec.~4.2 of the main paper, since our technique uses recursive decomposition, the inputs to the network beyond the first level would have at least one noise-free estimated image. However, our training involves noise for both the short exposure images. We showed results using this single-model approach in the main paper. As a variation to our single trained model, we also trained three different models (with the same architecture) for the three cases with two, one, and zero noisy images for the short-exposure inputs. This is explained in Fig.~\ref{fig:mult_models}.

\begin{figure*}
    \begin{center}
        \footnotesize
        \renewcommand{\tabcolsep}{0.5pt}
        \begin{tabular}{cccccc}
            \includegraphics[width=0.16\linewidth]{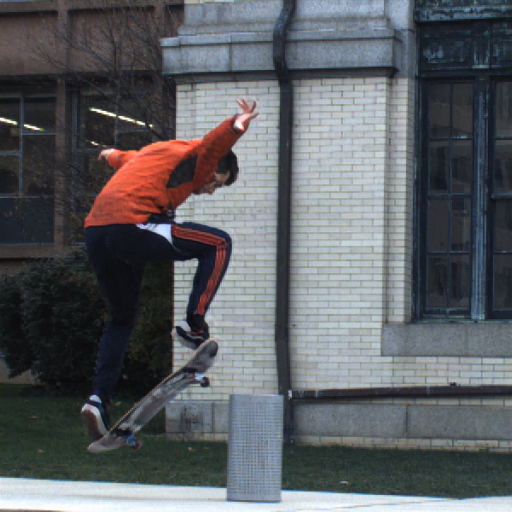}&
            \includegraphics[width=0.16\linewidth]{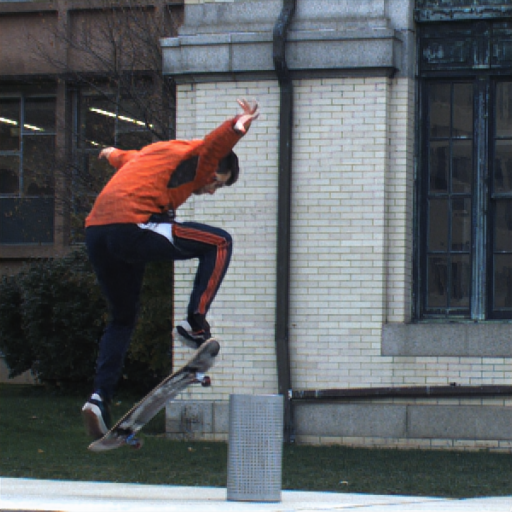}&
            \includegraphics[width=0.16\linewidth]{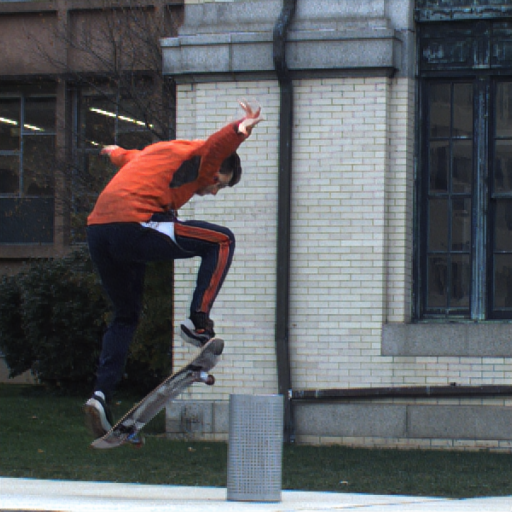}\hspace*{0.2cm}&
            \includegraphics[width=0.16\linewidth]{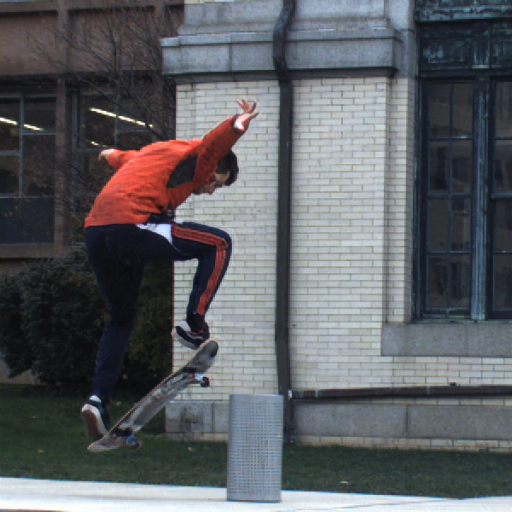}&
            \includegraphics[width=0.16\linewidth]{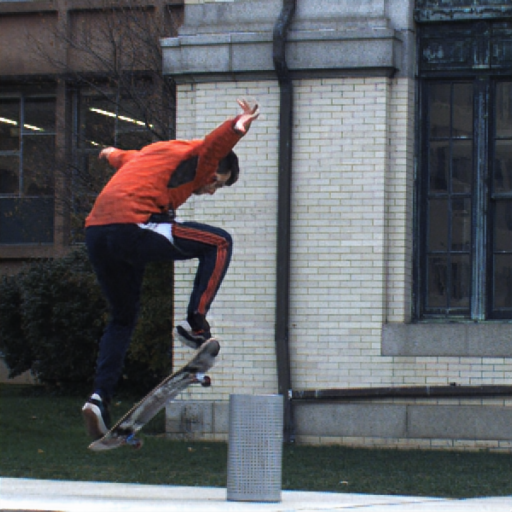}&
            \includegraphics[width=0.16\linewidth]{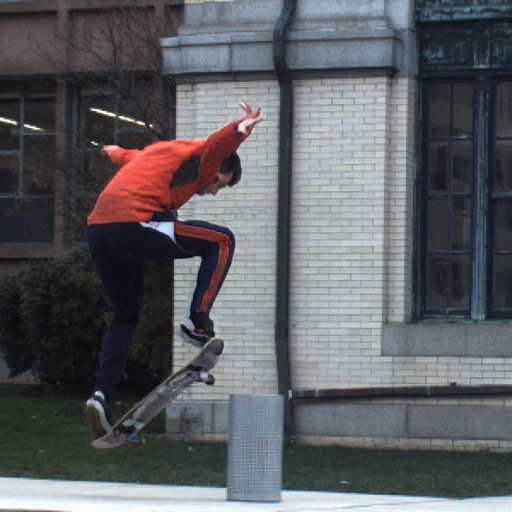}\\[-0.5cm]
            &&& {\color{yellow}Rel.~PSNR 43.83dB} & {\color{yellow}40.38 dB} & {\color{yellow}42.14 dB}\\[0.1cm]    
        \end{tabular}
        \begin{tabular}{cccccccccccc}
            \includegraphics[width=0.08\linewidth]{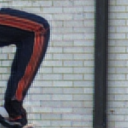}&
            \includegraphics[width=0.08\linewidth]{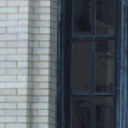}&
            \includegraphics[width=0.08\linewidth]{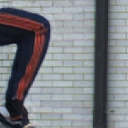}&
            \includegraphics[width=0.08\linewidth]{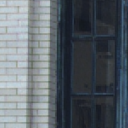}&
            \includegraphics[width=0.08\linewidth]{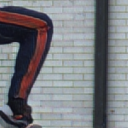}&
            \includegraphics[width=0.08\linewidth]{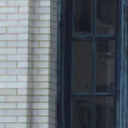}\hspace*{0.2cm}&
            \includegraphics[width=0.08\linewidth]{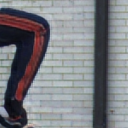}&
            \includegraphics[width=0.08\linewidth]{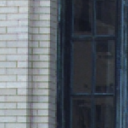}&
            \includegraphics[width=0.08\linewidth]{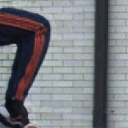}&
            \includegraphics[width=0.08\linewidth]{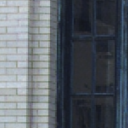}&
            \includegraphics[width=0.08\linewidth]{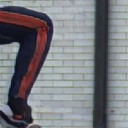}&
            \includegraphics[width=0.08\linewidth]{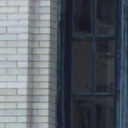}\\
            \multicolumn{2}{c}{$\widehat{I}_4$ (Model-0)} & \multicolumn{2}{c}{$\widehat{I}_6$ (Model-0)} & \multicolumn{2}{c}{$\widehat{I}_{12}$ (Model-0)} & 
            \multicolumn{2}{c}{$\widehat{I}_4$ (Model-1)} & \multicolumn{2}{c}{$\widehat{I}_6$ (Model-2)} & \multicolumn{2}{c}{$\widehat{I}_{12}$ (Model-1)} 
        \end{tabular}
        \begin{tabular}{cccccc}
            \includegraphics[width=0.16\linewidth]{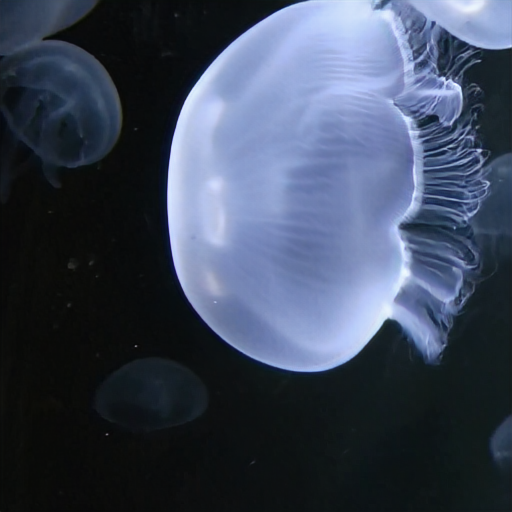}&
            \includegraphics[width=0.16\linewidth]{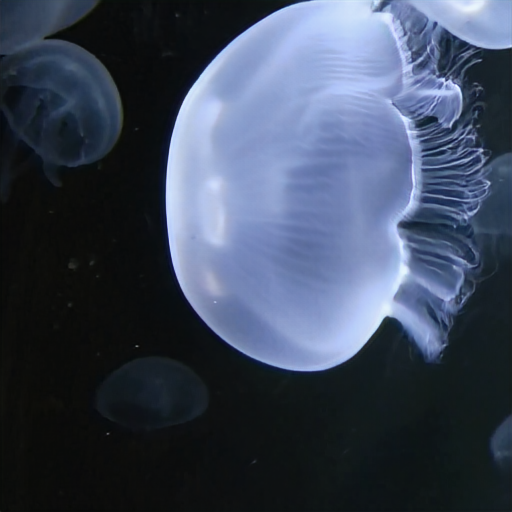}&
            \includegraphics[width=0.16\linewidth]{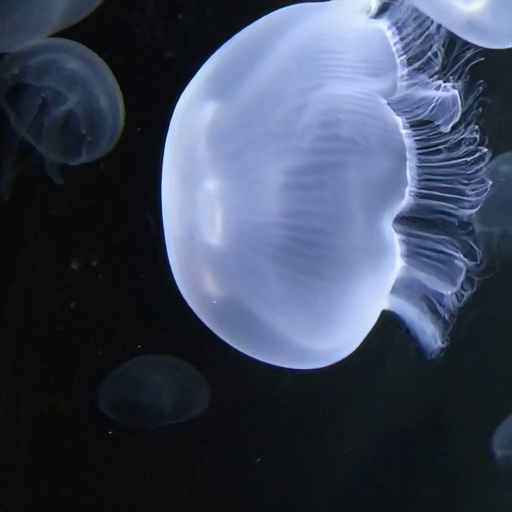}\hspace*{0.2cm}&
            \includegraphics[width=0.16\linewidth]{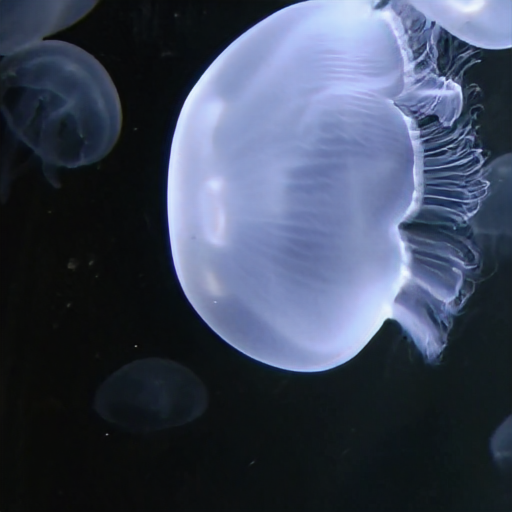}&
            \includegraphics[width=0.16\linewidth]{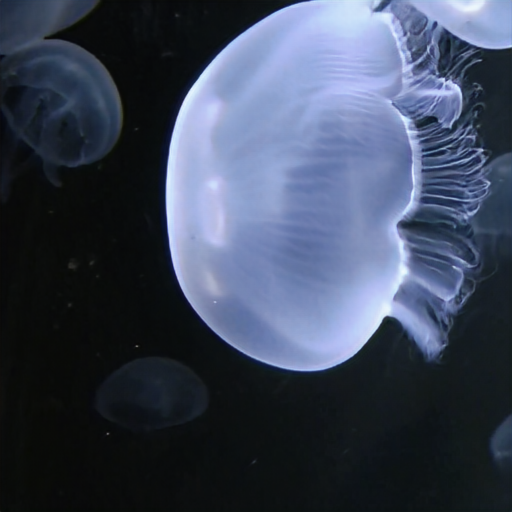}&
            \includegraphics[width=0.16\linewidth]{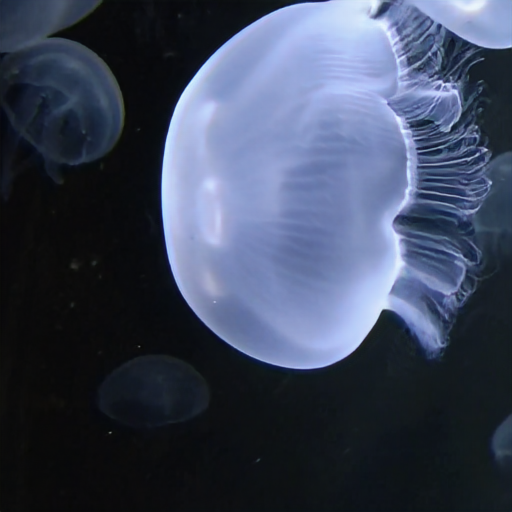}\\[-0.5cm]
            &&& {\color{yellow}Rel.~PSNR 43.62dB} & {\color{yellow}40.96 dB} & {\color{yellow}43.70 dB} \\[0.1cm]    
        \end{tabular}
        \begin{tabular}{cccccccccccc}
            \includegraphics[width=0.08\linewidth]{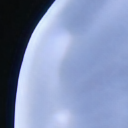}&
            \includegraphics[width=0.08\linewidth]{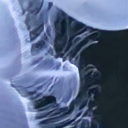}&
            \includegraphics[width=0.08\linewidth]{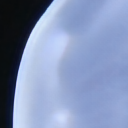}&
            \includegraphics[width=0.08\linewidth]{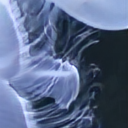}&
            \includegraphics[width=0.08\linewidth]{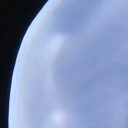}&
            \includegraphics[width=0.08\linewidth]{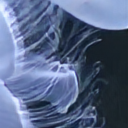}\hspace*{0.2cm}&
            \includegraphics[width=0.08\linewidth]{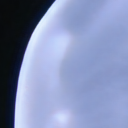}&
            \includegraphics[width=0.08\linewidth]{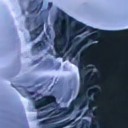}&
            \includegraphics[width=0.08\linewidth]{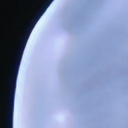}&
            \includegraphics[width=0.08\linewidth]{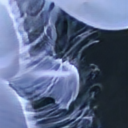}&
            \includegraphics[width=0.08\linewidth]{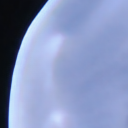}&
            \includegraphics[width=0.08\linewidth]{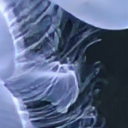}\\
            \multicolumn{2}{c}{$\widehat{I}_4$ (Model-0)} & \multicolumn{2}{c}{$\widehat{I}_6$ (Model-0)} & \multicolumn{2}{c}{$\widehat{I}_{12}$ (Model-0)} & 
            \multicolumn{2}{c}{$\widehat{I}_4$ (Model-1)} & \multicolumn{2}{c}{$\widehat{I}_6$ (Model-2)} & \multicolumn{2}{c}{$\widehat{I}_{12}$ (Model-1)} \\
            \multicolumn{6}{c}{(a) Single trained model} &  \multicolumn{6}{c}{(b) Three trained models}
        \end{tabular}
    \end{center}
    \caption{\textit{Comparison of single and three trained models approaches.} Both the approaches perform very similarly as shown in (a) and (b) for the estimation of three images of the sequence depicted in Fig.~\ref{fig:mult_models}. The relative PSNRs between the two approaches are quite high denoting this behavior.}
    \label{fig:mult_models_comp}
\end{figure*}

We observed no significant difference in our test results for these two approaches. This can be seen in Fig.~\ref{fig:mult_models_comp}. For four levels of decomposition as depicted in Fig.~\ref{fig:mult_models}, we show the estimated images $\widehat{I}_4$, $\widehat{I}_{12}$ inferred using Model-1 and $\widehat{I}_6$ using Model-2 in Fig.~\ref{fig:mult_models_comp}(b) for \textit{Skate} and \textit{Jellyfish} scenes. The corresponding estimated images using the single trained model approach inferred using Model-0 is shown in Fig.~\ref{fig:mult_models_comp}(a). We can see no observable difference. The relative PSNRs between the estimated images of the two approaches are on the high end showing that both the approaches performs very similarly, and hence, we used the single trained model approach for all our results in the main paper.

\section{Network Architecture}
\begin{figure}
    \begin{center}
        \hspace*{-0.75cm}\includegraphics[width=1.2\linewidth]{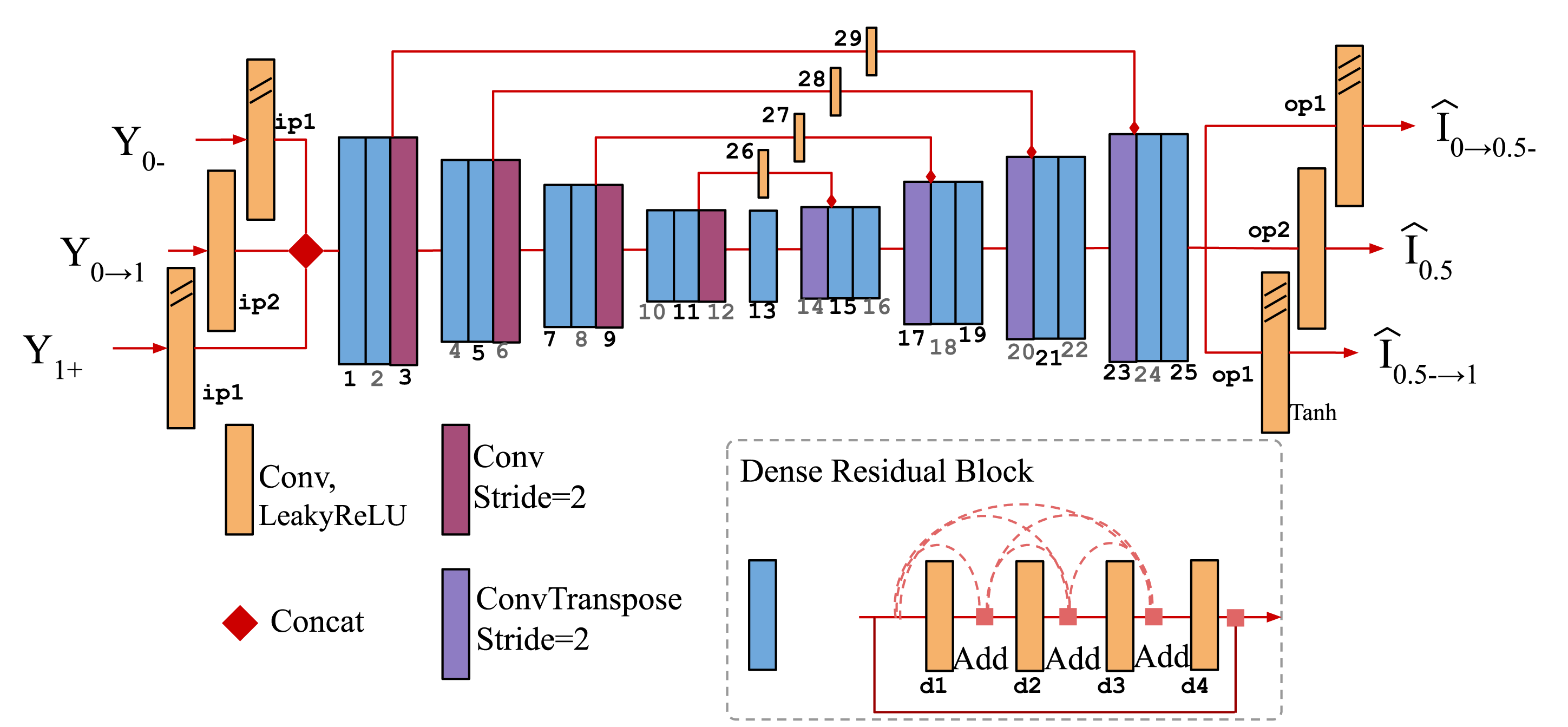}
    \end{center}        
    \caption{Network architecture.}
    \label{fig:supp_arch}
\end{figure}

The architecture of our neural network with layer numbering is shown in Fig.~\ref{fig:supp_arch}. The details of each layer, viz.~input/output channels, filter size, padding, and stride, are provided in Table \ref{tab:arch}.

\begin{table*}
    \caption{Layer details of our architecture in Fig.~\ref{fig:supp_arch}}
    \footnotesize
\begin{center}
    \begin{tabular}{cccc}
        \hline
        Layer & Type & chan-in $\rightarrow$ chan-out & filt, pad, stride\\ \hline
        ip1 & Conv & 3 $\rightarrow$ 16 & 7x7, 3, 1\\
        ip2 & Conv & 3 $\rightarrow$ 32 & 7x7, 3, 1\\
        \hdashline
        1 & ResB & 64 $\rightarrow$ 64 & 3x3, 1, 1\\
        2 & ResB & 64 $\rightarrow$ 64 & 3x3, 1, 1\\
        3 & Conv & 64 $\rightarrow$ 128 & 3x3, 1, {\color{downstride2} 2}\\
        \hdashline
        4 & ResB & 128 $\rightarrow$ 256 & 3x3, 1, 1\\
        5 & ResB & 128 $\rightarrow$ 256 & 3x3, 1, 1\\
        6 & Conv & 128 $\rightarrow$ 256 & 3x3, 1, {\color{downstride2} 2}\\
        \hdashline
        7 & ResB & 256 $\rightarrow$ 512 & 3x3, 1, 1\\
        8 & ResB & 256 $\rightarrow$ 512 & 3x3, 1, 1\\
        9 & Conv & 256 $\rightarrow$ 512 & 3x3, 1, {\color{downstride2} 2}\\
        \hdashline
        10 & ResB & 512 $\rightarrow$ 512 & 3x3, 1, 1\\
        11 & ResB & 512 $\rightarrow$ 512 & 3x3, 1, 1\\
        12 & Conv & 512 $\rightarrow$ 1024 & 3x3, 1, {\color{downstride2} 2}\\
        \hdashline
        13 & ResB & 1024 $\rightarrow$ 1024 & 3x3, 1, 1\\  \hline
    \end{tabular}\ \ \ \ 
    \begin{tabular}{cccc}
        &&&\\
        \hline
        Layer & Type & chan-in $\rightarrow$ chan-out & filt, pad, stride\\ \hline
        14 & ConvT & 1024 $\rightarrow$ 256 & 4x4, 1, {\color{upstride2} 2}\\
        15 & ResB & 512 $\rightarrow$ 512 & 3x3, 1, 1\\
        16 & ResB & 512 $\rightarrow$ 512 & 3x3, 1, 1\\
        \hdashline
        17 & ConvT & 512 $\rightarrow$ 128 & 4x4, 1, {\color{upstride2} 2}\\
        18 & ResB & 256 $\rightarrow$ 256 & 3x3, 1, 1\\
        19 & ResB & 256 $\rightarrow$ 256 & 3x3, 1, 1\\
        \hdashline
        20 & ConvT & 256 $\rightarrow$ 64 & 4x4, 1, {\color{upstride2} 2}\\
        21 & ResB & 128 $\rightarrow$ 128 & 3x3, 1, 1\\
        22 & ResB & 128 $\rightarrow$ 128 & 3x3, 1, 1\\
        \hdashline
        23 & ConvT & 128 $\rightarrow$ 32 & 4x4, 1, {\color{upstride2} 2}\\
        24 & ResB & 64 $\rightarrow$ 64 & 3x3, 1, 1\\
        25 & ResB & 64 $\rightarrow$ 64 & 3x3, 1, 1\\
        \hdashline
        op1 & Conv & 64 $\rightarrow$ 3 & 3x3, 1, 1\\
        op2 & Conv & 64 $\rightarrow$ 3 & 3x3, 1, 1\\
        \hdashline
        26 & Conv & 512 $\rightarrow$ 256 & 3x3, 1, 1\\
        27 & Conv & 256 $\rightarrow$ 128 & 3x3, 1, 1\\
        28 & Conv & 128 $\rightarrow$ 64 & 3x3, 1, 1\\
        29 & Conv & 64 $\rightarrow$ 32 & 3x3, 1, 1\\ \hline       
    \end{tabular}\\
    Conv - convolutional layer, ConvT - transpose convolutional layer~\cite{dumoulin:2016:guide}. \\
    All convolutional layers are followed by Leaky ReLU. Only op1 and op2 are followed by (tanh()+1)/2.\\ \ \\
    Structure of Dense Residual Block (ResB)\\
    \begin{tabular}{cccc}
        \hline
        Layer & Type & chan-in $\rightarrow$ chan-out & filt, pad, stride\\ \hline
        d1 & Conv & nchan $\rightarrow$ nchan & 3x3, 1, 1\\
        d2 & Conv & nchan $\rightarrow$ nchan & 3x3, 1, 1\\
        d3 & Conv & nchan $\rightarrow$ nchan & 3x3, 1, 1\\
        d4 & Conv & nchan $\rightarrow$ nchan & 3x3, 1, 1\\ \hline
    \end{tabular}
\end{center}
    \label{tab:arch}
\end{table*}

\end{document}